\documentclass[acmsmall,prologue,dvipsnames]{acmart}

\AtBeginDocument{%
  \providecommand\BibTeX{{%
    \normalfont B\kern-0.5em{\scshape i\kern-0.25em b}\kern-0.8em\TeX}}}

\usepackage{caption}
\usepackage{array}
\usepackage{multirow}
\usepackage{graphicx}
\usepackage{tablefootnote}
\usepackage[lofdepth,lotdepth]{subfig}
\newcommand{\symfootnote}[1]{%
\let\oldthefootnote=\thefootnote%
\stepcounter{mpfootnote}%
\addtocounter{footnote}{-1}%
\renewcommand{\thefootnote}{\fnsymbol{mpfootnote}}%
\footnote{#1}%
\let\thefootnote=\oldthefootnote%
}

\makeatletter
\DeclareRobustCommand{\cev}[1]{%
  {\mathpalette\do@cev{#1}}%
}
\newcommand{\do@cev}[2]{%
  \vbox{\offinterlineskip
    \sbox\z@{$\m@th#1 x$}%
    \ialign{##\cr
      \hidewidth\reflectbox{$\m@th#1\vec{}\mkern4mu$}\hidewidth\cr
      \noalign{\kern-\ht\z@}
      $\m@th#1#2$\cr
    }%
  }%
}
\makeatother

\definecolor{applegreen}{rgb}{0.25, 0.40, 0.0}

\setcopyright{acmcopyright}
\copyrightyear{2018}
\acmYear{2018}
\acmDOI{10.1145/1122445.1122456}

\acmJournal{JACM}
\acmVolume{37}
\acmNumber{4}
\acmArticle{111}
\acmMonth{8}

\begin{document}

\title{A Survey on Medical Document Summarization}


\author{Raghav Jain}
\affiliation{
  \institution{Indian Institute of Technology Patna}
  \city{Patna}
  \state{Bihar}
  \country{India}
}

\author{Anubhav Jangra}
\email{anubhav0603@gmail.com}
\orcid{0001-5571-6098}
\affiliation{
  \institution{Indian Institute of Technology Patna}
  \country{India}
}

\author{Sriparna Saha}
\affiliation{
  \institution{Indian Institute of Technology Patna}
  \city{Patna}
  \state{Bihar}
  \country{India}
}

\author{Adam Jatowt}
\affiliation{
  \institution{University of Innsbruck}
  \country{Austria}}

\renewcommand{\shortauthors}{Jain, et al.}

\begin{abstract} 
  
The internet has had a dramatic effect on the healthcare industry, allowing documents to be saved, shared, and managed digitally. This has made it easier to locate and share important data, improving patient care and providing more opportunities for medical studies. As there is so much data accessible to doctors and patients alike, summarizing it has become increasingly necessary - this has been supported through the introduction of deep learning and transformer-based networks, which have boosted the sector significantly in recent years. This paper gives a comprehensive survey of the current techniques and trends in medical summarization.
\end{abstract}

\ccsdesc[300]{Information systems~Similarity measures}
\ccsdesc[300]{Information systems~Information retrieval diversity}
\ccsdesc[500]{Information systems~Combination, fusion and federated search}
\ccsdesc[500]{Information systems~Language models}
\ccsdesc[300]{Information systems~Top-k retrieval in databases}
\ccsdesc[500]{Information systems~Speech / audio search}
\ccsdesc[500]{Information systems~Video search}
\ccsdesc[500]{Information systems~Image search}
\ccsdesc[300]{Information systems~Retrieval efficiency}
\ccsdesc[500]{Information systems~Summarization}
\ccsdesc[300]{Information systems~Information extraction}
\ccsdesc[300]{Computing methodologies~Neural networks}
\ccsdesc[300]{Computing methodologies~Supervised learning}
\ccsdesc[300]{Computing methodologies~Unsupervised learning}
\ccsdesc[500]{Computing methodologies~Natural language generation}
\ccsdesc[500]{Computing methodologies~Information extraction}

\keywords{summarization, clinical natural language processing, neural networks}

\maketitle

\section{Introduction} \label{sec:intro}

The internet has become a global phenomenon, connecting people all over the world and allowing for the exchange of information on a scale that was previously unimaginable. The rise of the internet and the corresponding digitization of many aspects of daily life has had a profound impact on society leading to information overload \cite{bontcheva2013social}. The sheer amount of information available today can be overwhelming. To combat this, individuals can use summarization techniques to distill the information down to its most essential points. The internet also had a profound impact on medical science. With the proliferation of online health tools, it is now easier than ever before to access medical information and resources \cite{november2012biomedical}. For example, individuals can easily search for medical information, research medical conditions and treatments, and find healthcare providers. Additionally, social media platforms have provided a platform for medical professionals to collaborate, share information, and discuss current medical topics. This has allowed medical professionals to quickly access the latest research, treatments, and developments in the field. Furthermore, online tools and platforms have enabled medical professionals to perform remote consultations with patients, providing more efficient and convenient healthcare services. There are a few reasons why summarization is important for medical documents. First, it allows for a quick overview of the document's content. This can be useful when trying to determine if the document is relevant to a particular topic of interest. Second, summarization can help to identify key points or ideas within a document. This can be valuable when trying to understand the main arguments or findings of a study. Finally, summarization can help to improve the readability of a document by reducing the amount of text that needs to be read. This application of summarization systems has the potential to reduce the burdens from medical workers who already are overburdened \cite{portoghese2014burnout}. \par
\begin{figure}[t]
\includegraphics[width=.7\textwidth]{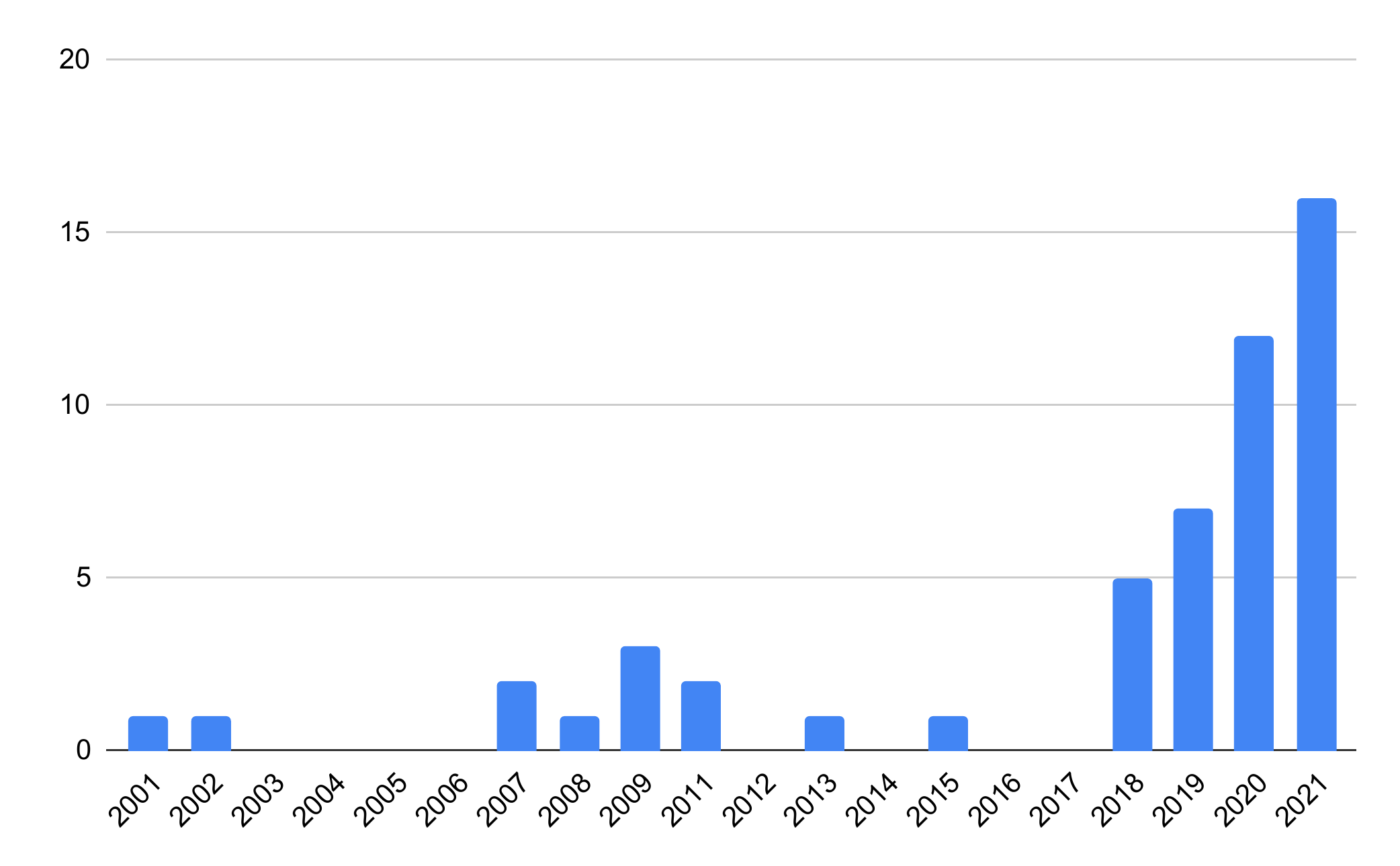}\centering
\caption{Trend in medical document  summarization research over last two decades. X-axis: year, Y-axis: \#papers on medical document  summarization published in each year. The growing number of papers in the recent 3 years suggests that there should be more coming in the next years.} \label{trend}
\end{figure}

Deep learning has been used in many other fields in addition to computer science \cite{sarker2021machine}, especially medical science. Deep learning can be used to diagnose diseases \cite{SHARMA202231}, predict patient outcomes \cite{xie2019use}, and even find new treatments \cite{bian2021generative}. In addition, deep learning can be used to analyze medical images, such as X-rays and MRI scans \cite{liu2021review}. There are many different applications for deep learning in medical science, and the potential benefits are huge. Deep learning could potentially revolutionize medicine, and make it more effective and efficient. One of such application area that leveraged AI and deep learning immensely is Clinical Natural Language Processing (NLP) \cite{wu2020deep} which can be defined as an interdisciplinary research field that involves the development of algorithms and systems to process natural language text from healthcare domains. It attempts to extract meaningful information from free-text clinical documents such as discharge summaries, clinical notes, and lab reports, in order to support clinical decision-making, clinical data mining, and other healthcare-related tasks. Clinical Natural language processing has gained a lot of popularity in the last few years due to its ability to enable better quality and cost-effective healthcare. The reason for this rise can be attributed to the release of large-scale datasets and different workshops \cite{bionlp-2022-biomedical} that are being organized to promote research in this area. Medical document summarization (MDS) can be considered as the subfield of clinical natural language processing. In the last few years, there have been various new advances in the field of medical document summarization, as illustrated in Figure \ref{trend}. This includes the introduction of new datasets, new methods for addressing the MDS problem, new challenges being organized in the community \cite{ben-abacha-etal-2021-overview,Nentidis_2021}, and the introduction of more suitable evaluation metrics. \par

At first, medical document summarization can be seen as a standard summarization problem only. However, after an extensive analysis, it can be observed that medical document summarization presents some unique and interesting challenges that may not be present in other domains because of the sensitivity of the medical domain and the complexity of the medical documents. These challenges expand the breadth of the problem, leading to a wider research scope for the task. The medical document summarization research field has seen a lot of progress in recent years. However, it is still disorganized compared to other areas in natural language processing. This lack of organization can make it difficult to gain an overall understanding of the research being conducted. There is currently no unified approach to this field, and research in this area continues to be fragmented. There is also lack of awareness about datasets available out there which leads to many works making their own datasets or not testing their approaches on other datasets. Additionally, resources that cover how to summarize medical documents and the types of medical documents available are not as readily available as they are in other fields. This restricts the exchange of ideas between different researchers. Furthermore, there has not been any standard evaluation set out there, so a conclusion on the best approach has not been possible. All of these factors work together to cause a lack of organization in the field of medical document summarization. This can make it difficult for researchers to keep track of the latest developments in the field. In addition, many of the papers published in this area are focused on specific applications of medical document summarization, such as health record summaries or research articles summaries. This makes it difficult to obtain an overall view of the state of the research field. These issues all together warrant the need for a survey on medical document summarization.

The rest of the paper is structured as follows. We first discuss related works and surveys in Section \ref{RW}. We formally define the MDS task in Section \ref{sec:MDS}. In
Section \ref{DD}, we provide an extensive categorization of existing works based on different types of medical tasks. In Section \ref{tax}, we provide an categorization of existing works based on the type of input, output, and techniques used. In Section \ref{sec:eval}, we discuss evaluation techniques devised for the evaluation of medical summaries. We then discuss the ethical concerns of deep learning while working in the medical domain in section \ref{EC}. We then provide a discussion based on all these in section \ref{Dis} followed by possibilities of future work in Section \ref{sec:future} and conclude our paper in Section \ref{sec:conc}.

\section{Related Works}\label{RW}
Recent advances in deep learning, typically, with the introduction of the transformers \cite{https://doi.org/10.48550/arxiv.1910.13461,https://doi.org/10.48550/arxiv.1910.10683,zhang2020pegasus}, have shown great success in text summarization leading to the study of different applications and domains within text summarization. With so much research going around summarization in the last few years, it is very difficult to be abreast with the current research and trends making it a necessity for researchers to work on survey papers around summarization so that community can catch up with this fast-moving area. A lot of application areas of summarization have been covered through different survey papers: Dialogue summarization \cite{feng2021survey}, Multimodal summarization \cite{https://doi.org/10.48550/arxiv.2109.05199}, Video summarization \cite{apostolidis2021video}, multi-view video summarization \cite{hussain2021comprehensive}, multi-document summarization \cite{ma2020multi}, multi-lingual summarization \cite{wang2022survey}, source code summarization \cite{zhang2022survey} etc. However, there is still a dearth of works focused on one of the most relevant use cases, i.e., medical document summarization. A thorough literature survey revealed that there is only one survey paper around medical document summarization \cite{afantenos2005summarization} that is from 2005. Apart from surveys about techniques for solving summarization, there is also a good amount of work around studying and comparing evaluation techniques \cite{https://doi.org/10.48550/arxiv.2007.12626, https://doi.org/10.48550/arxiv.2101.04840,bhandari-etal-2020-evaluating}.
\par
The medical domain has gained a lot of traction in the natural language processing community in the last few years. There are several surveys and comparative studies about the medical application of deep learning: medical imaging \cite{pandey2021comprehensive}, clinical NLP embeddings \cite{kalyan2020secnlp}, smart healthcare \cite{https://doi.org/10.48550/arxiv.2110.15803}, medical named entity recognition \cite{app11188319}, medical question answering \cite{10.1145/3490238} and medical dialogue \cite{valizadeh-parde-2022-ai} etc.\par

In this survey paper, we aim to present and discuss the research papers related to medical document summarization published during the period, 2015 January to 2022 March. Our contributions are as follows: (1) We have provided a new categorization of different medical document summarization subtasks based on the type of medical document detailing each type with its individual challenges and datasets to provide insights about specific documents to natural language processing researchers, (2) We have also classified existing works based on input, output, and {\bf SS:DID NOT GET method used type to provide a NLP point of view}, (3) We have also compiled different summary aspects that one must consider while evaluating medical summaries and then discussed new metrics developed in the community with their pros and cons.

\section{Medical Document Summarization task}
\label{sec:MDS}
In this section, we discuss different medical summarization tasks associated with different types of medical documents. Before diving into medical summarization, we broadly define the term automatic summarization. Automatic summarization can be defined as the task of computationally creating an abstract or a summary of original data while containing the relevant information and being consistent with the original data. Mathematically, automatic summarization is the task of producing the
\begin{equation}
\label{eq:1}
    Y_{summ}=f(D)
\end{equation}
such that $Length(Y_{summ})<Length(D)$ where $Y_{summ}$ is the output summary, $D$ is the input data and $f(.)$ is the the summarization function. When this input data $D$ contains medical information, the task of producing the summary, $Y_{summ}=f(D)$ is known as medical summarization task.\par
\begin{figure}[t]
\includegraphics[width=.6\textwidth]{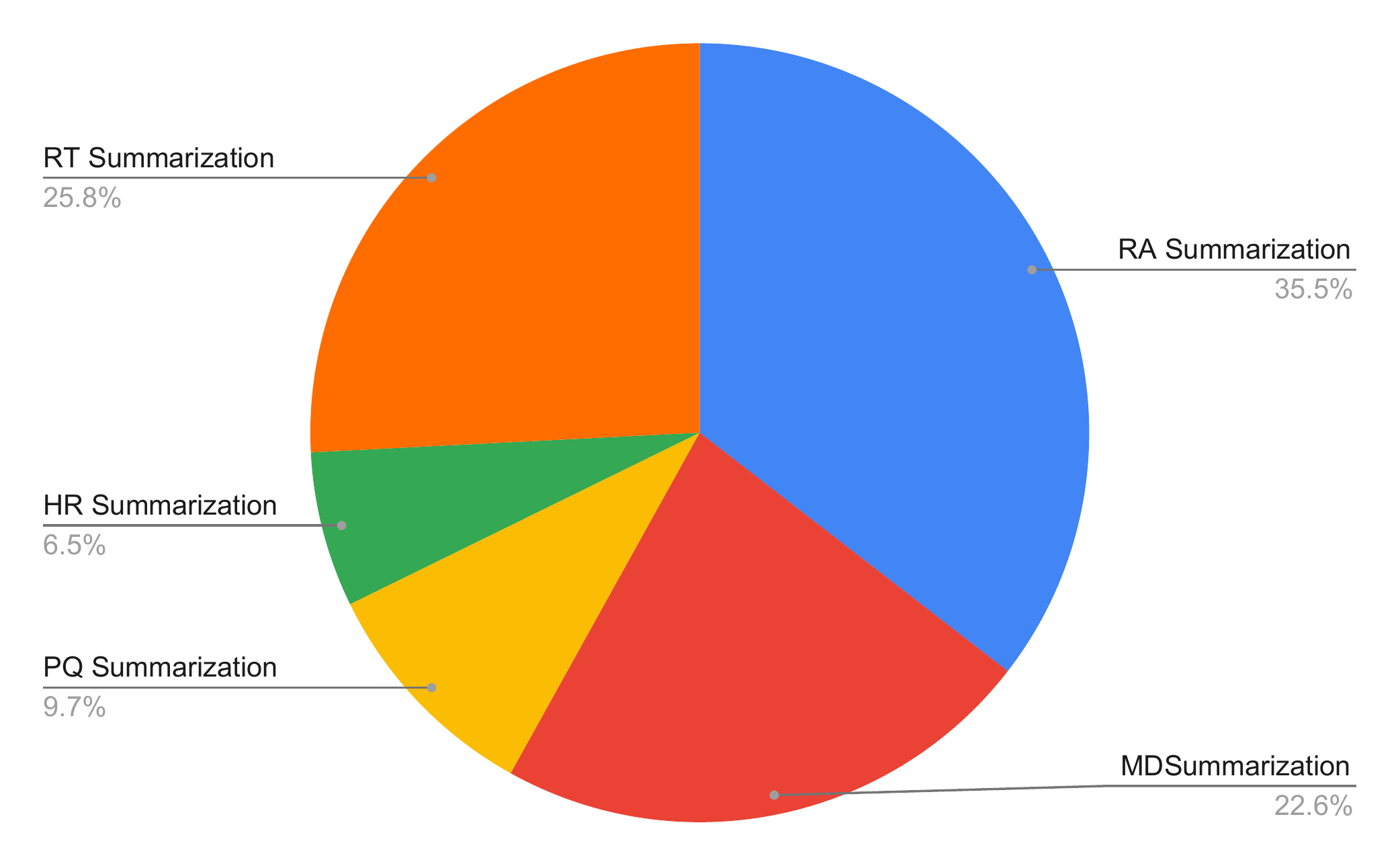}\centering
\caption{Illustration of distribution of existing work with respect to different MDS subtasks. Here, PQ: Patient Health Question, RA: Research Articles, HR: Health Records, MD: Medical Dialogue, RT: Report.} \label{MDStasks}
\end{figure}

Inspired by \cite{Afantenos2005SummarizationFM}, we can define medical information as the information and data concerning key concepts and techniques in the medical domain. Medical information may vary from core biomedical scientific concepts to a conversation between a doctor and a patient. Different types of medical information can be classified as follow:
\begin{enumerate}
    \item \textit{Biomedical Information:} It concerns with all the core medical science, diseases, health, and nutrition theories and concepts. 
    \item \textit{Clinical Information:} It refers to all the information generated from an interaction between a medical institution and a patient comprising of the patient's past medical records, medical reports generated from different tests, and the patient's hospital admission history, etc.
    \item \textit{Conversational Information:} It refers to all those semi-medical information that takes place on the web online or between doctor and patient such as medical queries and questions posted on different medical forums or a transcript of a conversation of a patient visiting a doctor.
\end{enumerate}
Based on these types of medical information and current literature, we can broadly divide the medical summarization task into different subtasks based on input data and documents associated with the medical domain (refer to Fig. \ref{MDStasks} for distribution of existing works) as follows:
\begin{itemize}
    \item \textbf{Report Summarization:} This refers to the summarization of notes or reports generated by a medical professional during the encounter with patients replacing input data $D$ in Equation \ref{eq:1} with clinical report. The main beneficiaries of summaries of the clinical reports are the medical practitioners as it will save their time and reduce their burden to go over the complete report.
\end{itemize}
\begin{itemize}
    \item \textbf{Health Record Summarization:} A health record refers to all the medical documents generated during different stages of the "journey" after a patient's admission to the hospital. Formally, Health Record Summarization can be defined as the multi-document summarization task where given a sequence of input documents $D=\{d\}$, we have to generate a summary, $Y_{summ}$. These health records are rich in medical information and summarizing these notes helps both doctor and patient to comprehend all the documents.
\end{itemize}
\begin{itemize}
    \item \textbf{Patient Health Question Summarization: } Consumer Health Question (CHQ) refers to the questions asked by patients to professional doctors and medical experts. Therefore, CHQ Summarization can be defined as the task of summarization of these medical questions by replacing input data $D$ in Equation \ref{eq:1} with CHQ. The main {\bf SS: DID NOT LIKE THE TERM audience for consumer health question summarization is medical doctors and physicians.}
\end{itemize}
\begin{itemize}
    \item \textbf{Medical Dialogue Summarization:} It can be defined as the process of summarizing a conversation over a digital platform or a physical encounter between a medical professional and a patient replacing the input data $D$ with Dialogue History. The users for Medical Dialogue Summarization are both patients and medical professionals as these summaries help to refer back to their interactions.
\end{itemize}
\begin{itemize}
    \item \textbf{Research Articles Summarization:} The aim of biomedical summarization systems is to summarize and present important and relevant medical facts and information from long articles to medical professionals. The main target {\bf I HAVE OBJECTION WITH THIS WORD audience} of this type of summarization are medical professionals and researchers.
\end{itemize}

\section{Deep dive in medical tasks}\label{DD}
\begin{figure}[t]
\includegraphics[width=.9\textwidth]{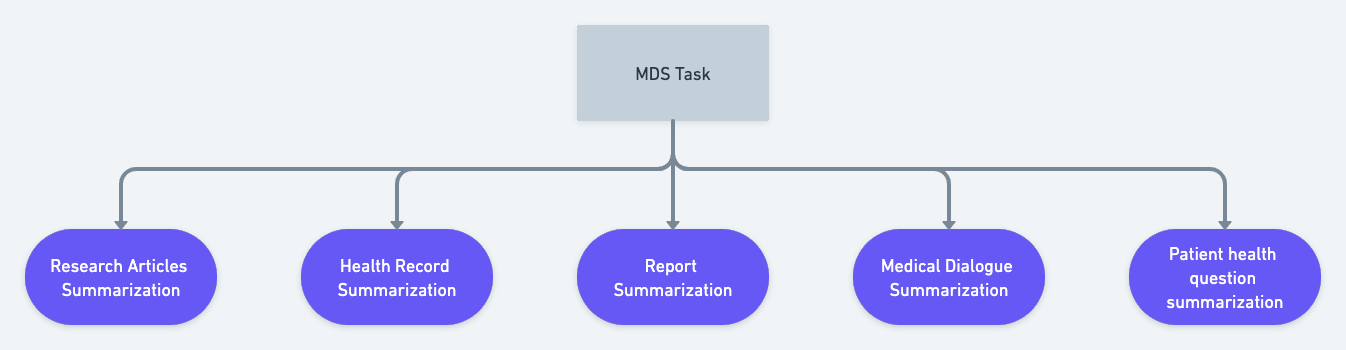}\centering
\caption{Visual representation of MDS subtasks.} \label{MDSvis}
\end{figure}

\begin{figure}[t]
\includegraphics[width=.7\textwidth]{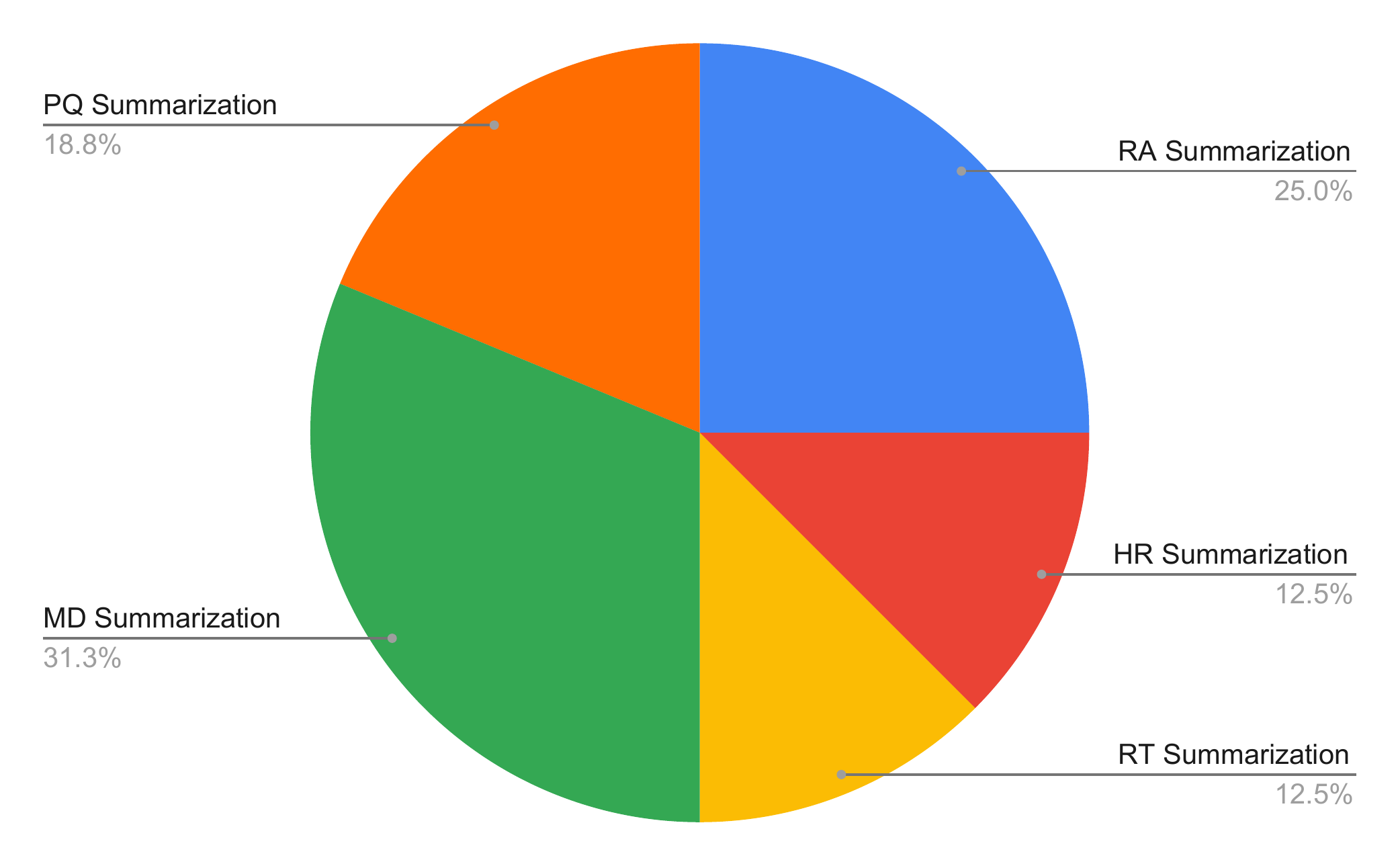}\centering
\caption{Illustration of dataset distribution with respect to different MDS subtasks. Here, PQ: Patient Health Question, RA: Research Articles, HR: Health Records, MD: Medical Dialogue, RT: Report.} \label{datasetdis}
\end{figure}

Before discussing the techniques and methods to solve MDS, we decided to first analyze and study the various types of MDS subtasks in detail. Since the MDS task is quite broad, it is important to categorize the works into different subtasks. For every subtask, we primarily discuss three issues: (1) Why this subtask is necessary, what type of document it deals with, and what is the structure of those documents?, (2)  What are the specific challenges that each subtask poses? (3) What are publicly available datasets (refer to Fig \ref{datasetdis} for distribution of datasets per subtask) for each subtask? We have also illustrated these
categorizations with a pictorial representation (Figure \ref{MDSvis}). We provided a comprehensive study of datasets in Table \ref{tab:datasets}.

\subsection{Research Articles Summarization}
Medical research articles are the most dominant form of spreading and sharing medical research advancements. With the advent of internet, the number of research articles in biomedical domain has grown exponentially. For example, PubMed which is the most popular repository for medical research articles contains more than 32 million articles \cite{https://doi.org/10.48550/arxiv.2110.11870}. Medical professionals face a lot of issues to keep themselves updated with the literature due to such a high rate of publications. This necessitates the need of automatic biomedical literature summarization systems which will help in reducing the burden of medical professionals. The aim of biomedical summarization systems is to summarize and present important and relevant medical facts and information from long articles to medical professionals. Most of the current works on research article summarization divide the articles on the following basis:
\begin{enumerate}
\item \textbf{Scientific medical research articles: } These are all the general medical articles which are centered around medical science, diseases, medical theories and concepts.
\item \textbf{COVID-19 specific research articles: } The emergence of the coronavirus has led to the explosion in the research articles discussing about the origin, symptoms, history and treatment of coronavirus. All the medical articles around COVID-19 come under this category.
\item \textbf{Health and Nutritional research articles: } These types of research articles contain medical information and statistics about food, nutrition and health \cite{https://doi.org/10.48550/arxiv.2103.11921}.

\item \textbf{Randomised Controlled Trials (RCTs): }Randomised Controlled Trials can be defined as the experiments and trials that aim to study the efficacy and success of new medical treatments and interventions \cite{hariton2018randomised}.
\end{enumerate}
We delineate the challenges, and dataset as follows:\par
\textbf{Challenges: } The primary challenge in summarizing medical articles is to handle the length of these medical documents as they are usually very long and also can contain multiple documents. As medical articles are published on a daily basis, there is a need to continuously update the existing summaries with the advent of new articles while retaining the old information \cite{https://doi.org/10.48550/arxiv.2007.03405, shah-etal-2021-nutri}. Shah et al. \cite{https://doi.org/10.48550/arxiv.2103.11921} also highlighted the issue of unfaithful summaries due to the limitations of deep learning models to comprehend relations (such as negation) between different entities. 
Other additional challenges include the low availability of some specific medical corpuses such as COVID-19 \cite{https://doi.org/10.48550/arxiv.2006.01997,PasqualiCRSJJ21} and esoteric medical terminology that may not be present in generic datasets. \par
\textbf{Datasets: } \textit{PubMed Open Access Subset}\footnote{\url{https://www.ncbi.nlm.nih.gov/pmc/tools/openftlist/}} is an online repository of PubMed scholarly articles which contains millions of journal articles from PubMed. Wang et al. \cite{wang2020cord} introduced a COVID-19 Open Research Dataset which includes 59,000 COVID-19 related research articles along with their corresponding summaries. Shah et al. \cite{https://doi.org/10.48550/arxiv.2103.11921} proposed a high-quality health and nutritional dataset which consists of 7,750 scientific abstracts as the document and human written summaries by doctors of those abstracts as output summary. DeYoung et al. \cite{deyoung-etal-2021-ms} developed a multi-document biomedical scientific literature summarization dataset, \textit{MSˆ2}, which contains 470k documents and 20K summaries from biomedical literature. Wallace et al. \cite{https://doi.org/10.48550/arxiv.2008.11293} also introduced a dataset for summarization of Randomized Control Trials (RCTs) derived from the Cochrane platform\footnote{\url{https://www.cochranelibrary.com/}}. \textit{BIOASQ} \cite{pubmed} is an open dataset containing 13 million PubMed research articles with their abstract as summaries of articles. 

\subsection{Health Record Summarization}
Electronic Health Records (EHR) \cite{ambinder2005electronic} are the digital documents that are used to document the encounter between medical professionals (physicians, doctors, nurses) and patients. EHR includes the complete course of the patient treatment from admission notes, doctor notes, nursing notes, and lab results to discharge notes. These records contain the complete the accounts of the journey of the patient in the hospital such as what happened to the patient, what treatment was given to the patient, and what are the future steps \cite{https://doi.org/10.48550/arxiv.2105.00816}.
Summarizing these numerous notes helps both doctor and patient to comprehend all the documents as these health records are rich in medical information. Thus making it a necessity to automate the summary generation of these health records for the medical professional \cite{care}. These summaries of EHR are commonly known as Discharge summaries \cite{https://doi.org/10.48550/arxiv.2104.13498}. This task of generating discharge summaries can be viewed as a multi-document summarization task consisting of structured (such as lab results) and unstructured (such as nursing notes) documents. We delineate the challenges, and dataset as follows:\par
\textbf{Challenges: } The existing works \cite{https://doi.org/10.48550/arxiv.2105.00816,https://doi.org/10.48550/arxiv.2104.13498} on summarizing electronic health records highlight the following challenges: (1) Size of the input: As electronic health records contain a sequence of long documents, it can exceed the maximum memory limit of current models. This will lead to truncation and information loss, (2) Evidence: As current deep learning frameworks are black boxes \cite{castelvecchi2016can}, medical professionals need a way to trace the origin of information present in summaries back to original documents, (3) Faithfulness: Medical domain is a highly critical area where there is no buffer for hallucination. Hallucination can be defined as the information generated by the model which is not present in the input document \cite{maynez2020faithfulness}, (4) Hybrid nature of input: Adams et al. \cite{https://doi.org/10.48550/arxiv.2105.00816} studied the EHR datasets and found that summaries in the dataset consist of both natures (Extraction and abstraction). This hybrid of extraction and abstraction warrants the use of models that can handle this transition between these two natures of summary, (5) Change in writing style: Adam et al. \cite{https://doi.org/10.48550/arxiv.2105.00816} also noted that discharge summaries also include a substantial amount of re-writing the content of EHRs as there is a transition in writing summaries from health records as EHR usually contains information in chronological order but discharge summaries are written from the patient's problems perspective, (6) Handling noisy reference summaries: Discharge summaries in EHR datasets usually either contains excessive information or a lack of critical medical information \cite{https://doi.org/10.48550/arxiv.2105.00816}. Such noisy reference summaries can harm the model's performance as the model will not be optimized for the right set of measures.

\textbf{Datasets: } Shing et al. \cite{https://doi.org/10.48550/arxiv.2104.13498} released a dataset of 6,000 encounters between patients and doctors. They derived the dataset from MIMIC-III clinical dataset \cite{johnson2016mimic}. MIMIC-III is a large open-source dataset that contains anonymous medical data of around 40,000 patients admitted to Beth Israel Deaconess Medical Center. Shing et al. \cite{https://doi.org/10.48550/arxiv.2104.13498} only selected those encounters that contain admission notes, ICU notes, radiology notes, echo notes, ECG notes, and
discharge summaries. Adams et al. \cite{https://doi.org/10.48550/arxiv.2105.00816} proposed a dataset called CLINSUM. CLINSUM contains medical records of 68,936 patients admitted to Columbia University Irving Medical Center from 2010 to 2014. They considered Brief Hospital Course (BHC) which is a mandatory section in discharge notes as the proxy reference summaries (Discharge summaries).
\subsection{Report Summarization}
Clinical reports and notes convey information about detailed medical observations and findings of a medical encounter between a medical professional and a patient. Summarizing these reports is a very critical process as these summaries are the primary source of information while reviewing patient medical history and contain critical medical information \cite{critical}. The most common type of clinical notes is radiology reports \cite{RR}. A radiology report is a medical document that contains the details of an imaging study (such as X-ray, MRI, etc). A Radiology report consists of three components: (1) \textbf{Background} section which contains the medical history of the patient, (2) \textbf{Findings} section which discusses the crucial observation and findings of the radiology study, and (3) \textbf{Impression} section which is a short summary of Findings section. \textbf{Impression} section is usually written by medical professionals which is a time-consuming process. The only aim of radiology report summarization is to automate the generation of this impression section. Most of the report summarization research revolves around the radiology domain because of the availability of large-scale datasets for radiology reports \cite{dataset1, https://doi.org/10.48550/arxiv.1901.07042}. However, all the existing work do not leverage the medical images in radiology reports to summarize the reports except for one work \cite{delbrouck-etal-2021-qiai}.
We delineate the challenges, and dataset as follows:

\textbf{Challenges: } Most of the current work in report summarization highlights the following two challenges: (1) Factual Correctness: Just like health record summarization, there is also no buffer hallucination and factual inconsistencies in report summarization, (2) Domain Specific Terminology: All the medical reports contain specific medical terminology that is usually not contained in normal/standard vocabulary and language models which warrants the use of external medical ontology and knowledge bases. Apart from this there is also need to understand the relationship between different medical terminology present in report.
\par

\textbf{Datasets: } MIMIC-CXR-JPG \cite{https://doi.org/10.48550/arxiv.1901.07042} is a large-scale freely available dataset of 377,110 chest x-rays associated with 227,827 imaging studies derived from the Beth Israel Deaconess Medical Center from 2011 to 2016. MIMIC-CXR-JPG is freely available on physionet\footnote{\url{https://physionet.org/}} which is an open repository of medical research data. Fushman et al. \cite{dataset1} released a dataset of 3,996 radiology reports from the Indiana Network for Patient Care and 8,121 associated images from the hospitals’ picture archiving systems. The images and reports were de-identified manually for ethical purposes.

\subsection{Medical Dialogue Summarization} Telemedicine has grown rapidly in the last few years with the aim of improving the efficiency of the healthcare system and reducing the workload of medical professionals. Telemedicine can be defined as the use of digital communication tools such as chatbots and chat interfaces with medical professionals to access the healthcare and medical services a person needs while practicing social distancing. With limited physical medical visits during the COVID-19 pandemic, telemedicine services and technologies have seen massive growth \cite{Mann2020COVID19TH}. Summarizing this conversation over telemedicine platforms is a significant step as it has multiple benefits: (1) Both doctor and patients have a record of their interaction, (2) the Doctor or patient can refer back to the conclusion or only to  important parts of the conversation, and (3) as a means of passing information to other medical professionals \cite{joshi-etal-2020-dr}. The goal of medical dialogue summarization is to extract relevant medical facts, information, symptoms, and diagnosis. The generated summary can be either in the form of structured notes \cite{krishna-etal-2021-generating, Liu2019TopicAwarePN} or unstructured summaries. The most common type of medical notes are SOAP notes (\textbf{S}ubjective information reported by patient, \textbf{O}bjective observations, \textbf{A}ssessment by medical professional and Future \textbf{P}lans) \cite{krishna-etal-2021-generating}. We delineate the challenges, and dataset as follows: \par
\textbf{Challenges: } The main challenge that restricted the research in medical Dialogue Summarization as compared to text summarization in the past was the lack of publically released datasets. Another challenge in dialogue summarization that is not present in normal text summarization is the dynamic flow of information which means that relevant medical information and facts are scattered across the entire conversation between both speakers. An ideal medical dialogue summarization system must understand the conversation flow to connect scattered utterances as these conversations can be asynchronous. While overcoming these challenges, the medical dialogue summarization system should capture all the relevant and important medical facts and information from the conversation such as symptoms, diagnosis, etc. \par

\textbf{Datasets: } Unlike normal text summarization, dialogue summarization is a much less explored area. But recently a lot of datasets have been developed to accelerate the research in medical dialogue summarization.
Krishna et al. \cite{krishna-etal-2021-generating} proposed a dataset of 6.5k clinical conversations along with SOAP notes as summaries. Song et al. \cite{song-etal-2020-summarizing} released a large dataset of 45k clinical conversations in the Chinese language. They map each conversation to two different summaries: (1) One summary of the medical problem reported by the patient, and (2) the Second summary of treatment suggested by a medical professional. Liu et al. \cite{Liu2019TopicAwarePN} proposed a structured summarization corpus of 100k dialogues where each dialogue is summarized into different predefined symptoms along with corresponding attributes. Joshi et al. \cite{joshi-etal-2020-dr} also created a corpus by extracting 25k medical conversations from a telemedicine platform and then hired doctors to annotate the conversations for corresponding summaries. Zhang et al \cite{https://doi.org/10.48550/arxiv.2109.12174} also build their own corpus of 1.3k medical conversations between doctors and patients.
\subsection{Patient health question summarization}
With the introduction of deep learning and transformer-based models, Question Answering (QA) systems has leveraged these techniques immensely. However, in the era of automation, there is still a dearth of works focused on one of the most relevant use cases, i.e., Medical QA. Recent surveys \cite{Jo2019ASR,FinneyRutten2019OnlineHI} also showed an increase in patients seeking medical information on the web these days. However, the main challenge impeding the research of Medical QA is the complexity of the questions posted by consumers/patients, both in the length of the question and the information contained in that question. These questions usually contain redundant and irrelevant information that is not required to answer the question by doctors or physicians. \citet{10.1093/jamia/ocw024} highlighted this issue of irrelevant information in medical QA by showing that questions posted by patients are filled with more background information as compared to professional questions. \citet{Mrini2021AGS} also reports that patients often use medical terms and vocabulary different from 
\begin{table}[H]  \centering
\scriptsize
\caption{\textbf{A study on datasets available for medical document summarization.} `T' stands for English text, 'SD' stands for Single Document, 'MD' stands for Multi-Document, 'SS' stands for structured summary, 'MS' stands for multiple summaries, `TC' stands for Chinese text, 'TE' stands for text (extractive) summary, `TA' stands for text (abstractive) summary, `I' stands for images.}\label{tab:datasets}
\renewcommand{\arraystretch}{2}
\begin{tabular}{|l|p{0.1\textwidth}|p{0.1\textwidth}|p{0.3\textwidth}|p{0.12\textwidth}|}
\hline
\textbf{ID \& Paper} & \textbf{Type of Input data} & \textbf{Type of output data} & \textbf{Data Statistics} & \textbf{MDS subtask}  \\
\hline
\#1: Wang et al. \cite{wang2020cord} & T, SD & TA & 59,000 COVID-19 Research articles & Research Articles Summarization \\\hline
\#2: Shah et al. \cite{https://doi.org/10.48550/arxiv.2103.11921} & T, SD & TA & 7,750 scientific articles & Research Articles Summarization \\\hline
\#3: DeYoung et al. \cite{deyoung-etal-2021-ms} & T, MD & TA & 470k documents
and 20K summaries from biomedical literature & Research Articles Summarization \\\hline
\#4: \citet{pubmed} & T, SD & TA & 13 million PubMed research articles  & Research Articles Summarization \\\hline
\#5: Shing et al. \cite{https://doi.org/10.48550/arxiv.2104.13498} & T, MD & TA & dataset of 6,000 encounters between patients and doctors & Health Record Summarization \\\hline
\#6: Adams et al. \cite{https://doi.org/10.48550/arxiv.2105.00816} & T, MD & TA & medical records of 68,936 patients & Health Record Summarization \\\hline
\#7: \citet{https://doi.org/10.48550/arxiv.1901.07042} & T, I & TA & Large scale freely available dataset of summaries of  377,110 chest x-ray s & Report Summarization \\\hline
\#8: Fushman et al. \cite{dataset1} & T, I & TA &  3,996 radiology reports from the
Indiana Network for Patient Care and 8,121 associated images from the hospitals’ picture archiving
systems &Report Summarization  \\\hline
\#9: Krishna et al. \cite{krishna-etal-2021-generating}  & T,SD & TA, SS & dataset of 6.5k clinical conversations along with
SOAP notes as summaries &Medical Dialogue Summarization  \\\hline
\#10: Song et al. \cite{song-etal-2020-summarizing} & TC, SD & TE, MS &  dataset of 45k clinical conversations & Medical Dialogue Summarization \\\hline
\#11: Liu et al. \cite{Liu2019TopicAwarePN} & T, SD & TA, SS & structured summarization corpus of 100k dialogues &Medical Dialogue Summarization   \\\hline
\#12: Joshi et al. \cite{joshi-etal-2020-dr} & T, SD& TA & 25k medical conversations from a
telemedicine platform & Medical Dialogue Summarization  \\\hline
\#13:  Zhang et al \cite{https://doi.org/10.48550/arxiv.2109.12174}  & T, SD & TA & 1.3k medical conversations between doctors and patients& Medical Dialogue Summarization \\\hline
\#14: \citet{Abacha2019OnTS} & T, SD& TA & 1000 health questions along with their corresponding
manually written gold standard summaries & Patient health question summarization\\\hline
\#15: \citet{Mrini2021JointSO} & T, SD & TA & 2,26,405 question summary pair from HealthCareMagic platform & Patient health question summarization \\\hline
\#16: \citet{Mrini2021JointSO} & T, SD & TA & 31,062 question summary pairs from  iCliniq. platform & Patient health question summarization\\\hline
\end{tabular}
\end{table}   
professional doctors. These issues necessitate the need for the summarization of consumer health questions (CHQ). \citet{MQA} also showed that summarization of CHQs

improves the performance of QA systems by a margin of 58\%. We delineate the challenges, and dataset as follows: \par
\textbf{Challenges: } Like any other domain-specific summarization, one of the main challenges in CHQ summarization is capturing all the relevant medical domain terminologies and entities in the generated summary. A summarization system should also understand the relationship between different medical entities to make summaries more semantically consistent. Apart from these domain-specific challenges, there are also challenges that are distinct to the summarization of questions. The system should understand the type and focus of the question to make summary more meaningful. It should also capture all the sub-questions present in the original consumer question.\par

\textbf{Datasets: } Recently, a few datasets have been proposed to aid the progress in CHQ summarization task. \citet{Abacha2019OnTS} proposed a dataset called MeQSum. MeQSum is a medical question summarization dataset that contains 1000 health questions along with their corresponding manually written gold standard summaries. \citet{Abacha2019OnTS} also showed two data augmentation techniques to increase the dataset size to 5,155 and 8,014 question summary pairs respectively. \citet{Mrini2021JointSO} also created two CHQ datasets, extracting from a large-scale medical dialogue dataset MedDialog \cite{zeng-etal-2020-meddialog} named HealthCareMagic and iCliniq. HealthCareMagic contains a total of 2,26,405 question summary pairs whereas iCliniq contains 31,062 question summary pairs.

\section{Organization of existing work} \label{tax}
\begin{figure}[t]
\includegraphics[width=\textwidth]{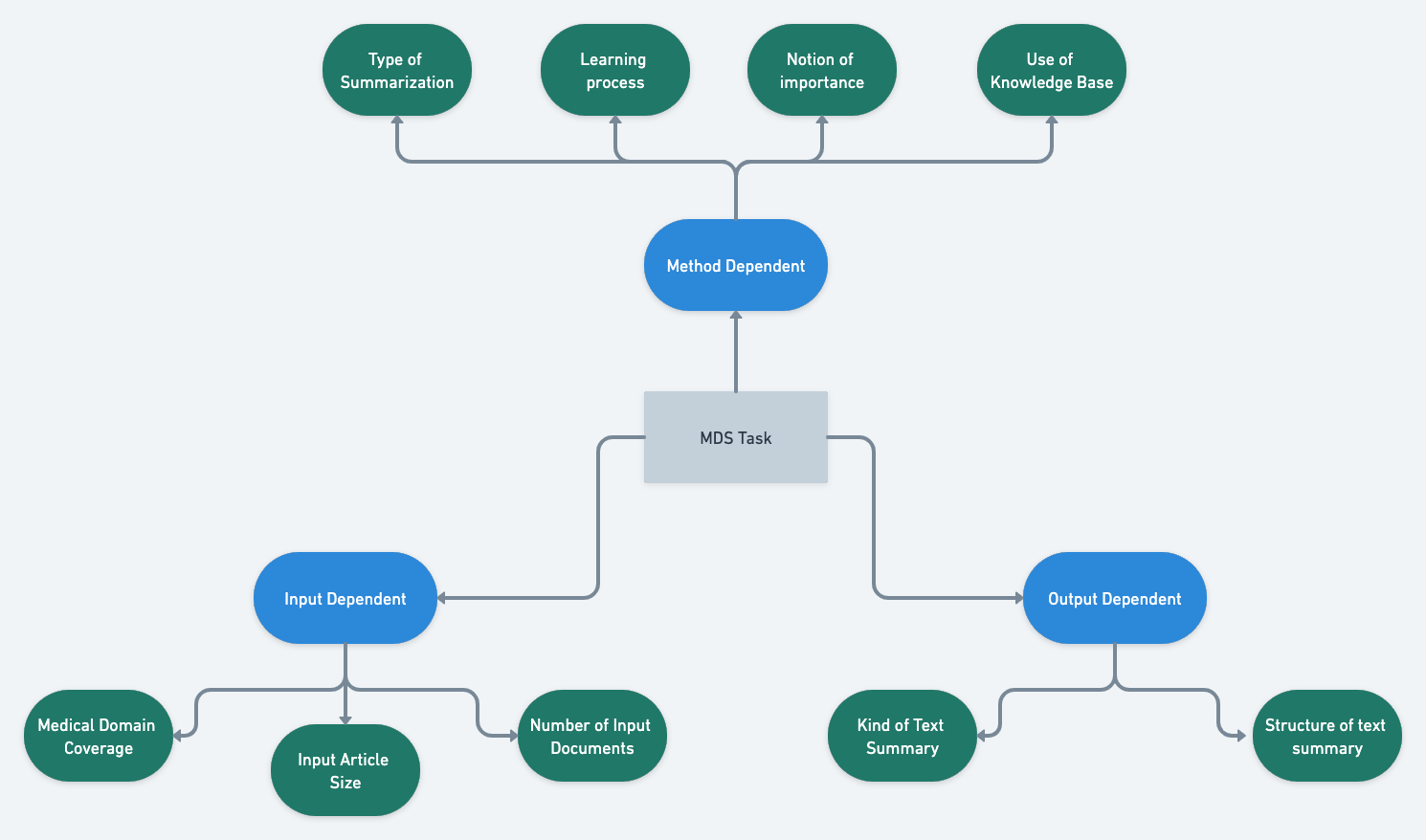}\centering
\caption{Visual representation of proposed taxonomy based on input type, output type and method type.} \label{taxonomy}
\end{figure}

There have been many attempts to solve the medical document summarization task, so it is important to categorize the existing works and methods to get a clear picture of the task and understand the current trends. We categorize the prior works into three broad categories, depending upon the variations in input, output, and techniques used. We have also illustrated these categorizations with a pictorial representation (Figure \ref{taxonomy}) and provided a comprehensive study in Table \ref{tab:categ} (note that if some classifications are not marked in the table, then either the information about that category was not present, or is not applicable.).
\subsection{On the basis of Input}
A summarization task is highly driven by the kind of input it is given. An existing
work can be distinguished from others on the basis of input in the following categories:\par
\textbf{Medical Domain Coverage: } Depending upon the extent of medical domain coverage, we can classify existing works as medical domain-specific or medical generic. The approach to summarize a domain-specific input can differ from the generic
input greatly since feature extraction in the former can be very particular in nature and can have pre-requisites or specific standards to follow while generating summary. Most of the biomedical documents summarization \cite{https://doi.org/10.48550/arxiv.2008.11293, https://doi.org/10.48550/arxiv.2012.12573, deyoung-etal-2021-ms, https://doi.org/10.48550/arxiv.2110.11870,du2020biomedical,kedzie2018content} are generic in nature since these articles contain information about almost all the medical
domains; whereas Covid-19 research articles summarization \cite{https://doi.org/10.48550/arxiv.2006.01997, cai2022covidsum,https://doi.org/10.48550/arxiv.2007.03405}, Nutrition Articles summarization \cite{shah-etal-2021-nutri}, Radiology report summarization \cite{zhang2019optimizing, hu2021word} are  some examples of medical domain-specific MDS tasks.
\par
\textbf{Input Article Size: } Since most of the work discussed in this survey has text modality as the input, the size of the text document in input can also be one way of categorizing all the related works. The summarization strategies might be different depending upon whether the textual input is a short single passage \cite{Abacha2019OnTS, CHQ, Mrini2021JointSO} or a multi-passage \cite{zeng-etal-2020-meddialog, joshi-etal-2020-dr, delbrouck-etal-2021-qiai, https://doi.org/10.48550/arxiv.2104.13498}. Patient health question summarization come under the single passage category and the rest of the MDS subtasks are multi-passage.
\par
\textbf{Number of Input Documents: } One way to categorize the existing works is by the number of text documents in the input i.e. Multi-document summarization that is the task of generating a summary that consists of information from multiple documents or single-document summarization that is the task of generating a summary from a single document. In MDS, multi-document summarization mainly includes Health Record Summarization \cite{https://doi.org/10.48550/arxiv.2104.13498, https://doi.org/10.48550/arxiv.2105.00816} and single-document summarization includes all other subtasks \cite{Abacha2019OnTS, CHQ, Mrini2021JointSO, zeng-etal-2020-meddialog, joshi-etal-2020-dr, delbrouck-etal-2021-qiai}.
\par

\subsection{On the basis of Output}
We can group the existing works on the basis of type of output into the following categories:\par
\textbf{Kind of text summary: }Most text summarization works fall into two categories: extractive and abstractive. Extractive summaries are those that contain information from the text directly, while abstractive summaries are those that involve creating a new summary that is focused on the overall meaning of the text. Depending on this, we can also classify MDS tasks as either extractive \cite{du2020biomedical, song2020summarizing} or abstractive \cite{https://doi.org/10.48550/arxiv.2006.01997, cai2022covidsum,https://doi.org/10.48550/arxiv.2007.03405,Abacha2019OnTS, CHQ, Mrini2021JointSO, zeng-etal-2020-meddialog, joshi-etal-2020-dr, delbrouck-etal-2021-qiai}.
\par
\textbf{Structure of text summary: } Structured text summary is a summary of a text that includes the most important information from the text in a well-organized format. An unstructured text summary is a summary of a text that includes some important information from the text but is not as well-organized as a structured text summary. On this basis, we can categorize current works into two categories: Structured MDS which generates a medical summary in a specific format such as SOAP notes \cite{krishna-etal-2021-generating} or symptoms with attributes \cite{Liu2019TopicAwarePN} and Unstructured MDS that generates medical summary in a normal generic paragraph fashion \cite{https://doi.org/10.48550/arxiv.2104.13498, du2020biomedical}.
\par

\begin{table}[tp]  \centering
\scriptsize
\caption{\textbf{Comprehensive list of works that uses different deep learning techniques to generate output summary.}}\label{tab:DL}
\renewcommand{\arraystretch}{2}
\begin{tabular}{|p{0.2\textwidth}|p{0.6\textwidth}|}
\hline
\textbf{Deep Learning Technique} & \textbf{Works using this technique}\\
\hline
Seq2seq Learning Framework  & \citet{zhang2018learning}, \citet{zhang2019optimizing}, \citet{macavaney2019ontology}, \citet{sotudeh2020attend}, \citet{shah2021nutri}, \citet{https://doi.org/10.48550/arxiv.2110.11870}, \citet{kedzie2018content}, \citet{song-etal-2020-summarizing}, \citet{Liu2019TopicAwarePN}
\\\hline

Transformer based Networks  & \citet{https://doi.org/10.48550/arxiv.2104.13498}, \citet{kondadadi2021optum},\citet{https://doi.org/10.48550/arxiv.2006.01997}, \citet{https://doi.org/10.48550/arxiv.2007.03405}, \citet{https://doi.org/10.48550/arxiv.2008.11293}, \citet{https://doi.org/10.48550/arxiv.2012.12573}, \citet{deyoung-etal-2021-ms}, \citet{du2020biomedical}, \citet{krishna-etal-2021-generating}, \citet{enarvi2020generating}, \citet{chintagunta-etal-2021-medically}, \citet{joshi-etal-2020-dr} \\\hline

Graph Deep Learning  & \citet{hu2021word}, \citet{cai2022covidsum} \\\hline

\end{tabular}
\end{table}
\subsection{On the basis of Method}
Different ways of solving the MDS problem have been developed, which can be divided into categories according to the method used as follows: \par
\textbf{Type of Summarization: }Broadly, existing approaches to MDS can be classified into the following: Extractive MDS \cite{du2020biomedical, song2020summarizing} that involves the use of techniques that copies text from the input source itself; Abstractive MDS \cite{https://doi.org/10.48550/arxiv.2006.01997, cai2022covidsum,https://doi.org/10.48550/arxiv.2007.03405,Abacha2019OnTS, CHQ, Mrini2021JointSO, zeng-etal-2020-meddialog, joshi-etal-2020-dr, delbrouck-etal-2021-qiai} that involve rewriting the input document completely into a compressed form; and Hybrid (Extract-then-Abstract) \cite{https://doi.org/10.48550/arxiv.2101.04840} MDS that first extracts salient and relevant information then paraphrases and compresses it to form final summary. The motivation for extract-then-abstract is that extractive models are better at being faithful to the source, but abstractive models are better at producing coherent summaries; thus combining the best from both worlds to produce a faithful and fluent summary.
\par
\begin{table}
 \setlength\extrarowheight{1pt}
 \centering
 \scriptsize
 \caption{\textbf{Comprehensive study of existing work using the proposed taxonomy (refer to Section \ref{tax})}.}
 \label{tab:categ}
 \vspace{-3mm}
 \rotatebox{90}{
 \begin{tabular}{|l*{21}{|c}|}
 \hline
   & \multicolumn{6}{c|}{\textbf{Input Based}} & \multicolumn{4}{c|}{\textbf{Output Based}} & \multicolumn{11}{c|}{\textbf{Method Based}} \\ \cline{2-22}
  
  \multirow{2}{*}{\textbf{\hspace{10mm} Papers}} & \multicolumn{2}{c|}{\textbf{MDC}} & \multicolumn{2}{c|}{\textbf{IAS}} & \multicolumn{2}{c|}{\textbf{IDN}} & \multicolumn{2}{c|}{\textbf{KTS}} & \multicolumn{2}{c|}{\textbf{STS}} & \multicolumn{3}{c|}{\textbf{TS}} & \multicolumn{3}{c|}{\textbf{LP}} & \multicolumn{3}{c|}{\textbf{NI}} & \multicolumn{2}{c|}{\textbf{KB}} \\ \cline{2-22}
  
   & \rotatebox{270}{Specific Medical Domain \phantom{.}} & \rotatebox{270}{Generic\phantom{.}} &  \rotatebox{270}{Single-Passage\phantom{.}} & \rotatebox{270}{Multi-Passage\phantom{.}} & \rotatebox{270}{Single-doc\phantom{.}} &   \rotatebox{270}{Multi-doc\phantom{.}} &  \rotatebox{270}{Extractive\phantom{.}} & \rotatebox{270}{Abstractive\phantom{.}} & \rotatebox{270}{Structured summary\phantom{.}} & \rotatebox{270}{Unstructured summary\phantom{.}} & \rotatebox{270}{Abstractive\phantom{.}} & \rotatebox{270}{Extractiive\phantom{.}} &
   \rotatebox{270}{Hybrid\phantom{.}} &
   \rotatebox{270}{Rule Based\phantom{.}} & \rotatebox{270}{Machine learning\phantom{.}} & \rotatebox{270}{Deep learning\phantom{.}} & \rotatebox{270}{Consistency\phantom{.}} & \rotatebox{270}{Copy mechanism\phantom{.}} & \rotatebox{270}{Other\phantom{.}} & \rotatebox{270}{KB used\phantom{.}} & \rotatebox{270}{KB not used\phantom{.}}  \\ \hline

  \citet{https://doi.org/10.48550/arxiv.2104.13498} &  &\textcolor{applegreen}{\textbf{\checkmark}}   &  &\textcolor{applegreen}{\textbf{\checkmark}} &  & \textcolor{applegreen}{\textbf{\checkmark}} &
  & \textcolor{applegreen}{\textbf{\checkmark}} & & \textcolor{applegreen}{\textbf{\checkmark}} &  &  &\textcolor{applegreen}{\textbf{\checkmark}} &  & & \textcolor{applegreen}{\textbf{\checkmark}}& && \textcolor{applegreen}{\textbf{\checkmark}} &   &\textcolor{applegreen}{\textbf{\checkmark}} \\ \hline
  
    \citet{https://doi.org/10.48550/arxiv.2006.01997} &\textcolor{applegreen}{\textbf{\checkmark}}  &   &  &\textcolor{applegreen}{\textbf{\checkmark}} &  \textcolor{applegreen}{\textbf{\checkmark}}&  &
  & \textcolor{applegreen}{\textbf{\checkmark}} & & \textcolor{applegreen}{\textbf{\checkmark}} &\textcolor{applegreen}{\textbf{\checkmark}}  &  & &  & & \textcolor{applegreen}{\textbf{\checkmark}}& && \textcolor{applegreen}{\textbf{\checkmark}} &   &\textcolor{applegreen}{\textbf{\checkmark}} \\ \hline

        \citet{sarkar2011using} &   &\textcolor{applegreen}{\textbf{\checkmark}}  &  & \textcolor{applegreen}{\textbf{\checkmark}}&  \textcolor{applegreen}{\textbf{\checkmark}}&  &
  \textcolor{applegreen}{\textbf{\checkmark}} & &  &\textcolor{applegreen}{\textbf{\checkmark}} &  &\textcolor{applegreen}{\textbf{\checkmark}}  & &  &\textcolor{applegreen}{\textbf{\checkmark}} & & &&\textcolor{applegreen}{\textbf{\checkmark}}  &    &\textcolor{applegreen}{\textbf{\checkmark}}\\ 
  \hline
  
      \citet{cai2022covidsum} &\textcolor{applegreen}{\textbf{\checkmark}}  &   &  &\textcolor{applegreen}{\textbf{\checkmark}} &  \textcolor{applegreen}{\textbf{\checkmark}}&  &
  & \textcolor{applegreen}{\textbf{\checkmark}} & & \textcolor{applegreen}{\textbf{\checkmark}} &\textcolor{applegreen}{\textbf{\checkmark}}  &  & &  & & \textcolor{applegreen}{\textbf{\checkmark}}& && \textcolor{applegreen}{\textbf{\checkmark}} &  \textcolor{applegreen}{\textbf{\checkmark}} & \\ \hline
  
        \citet{https://doi.org/10.48550/arxiv.2007.03405} &\textcolor{applegreen}{\textbf{\checkmark}}  &   &  &\textcolor{applegreen}{\textbf{\checkmark}} &  \textcolor{applegreen}{\textbf{\checkmark}}&  &
  \textcolor{applegreen}{\textbf{\checkmark}}&  & & \textcolor{applegreen}{\textbf{\checkmark}} &  &\textcolor{applegreen}{\textbf{\checkmark}}  & &  & & \textcolor{applegreen}{\textbf{\checkmark}}& && \textcolor{applegreen}{\textbf{\checkmark}} &   &\textcolor{applegreen}{\textbf{\checkmark}} \\ \hline
          \citet{https://doi.org/10.48550/arxiv.2008.11293} &  &\textcolor{applegreen}{\textbf{\checkmark}}   &  &\textcolor{applegreen}{\textbf{\checkmark}} &  &\textcolor{applegreen}{\textbf{\checkmark}}  &
  &\textcolor{applegreen}{\textbf{\checkmark}}  & & \textcolor{applegreen}{\textbf{\checkmark}} &  \textcolor{applegreen}{\textbf{\checkmark}}&  & &  & & \textcolor{applegreen}{\textbf{\checkmark}}& && \textcolor{applegreen}{\textbf{\checkmark}} &   \textcolor{applegreen}{\textbf{\checkmark}}& \\ \hline
            \citet{https://doi.org/10.48550/arxiv.2103.11921} &  \textcolor{applegreen}{\textbf{\checkmark}} &  &  &\textcolor{applegreen}{\textbf{\checkmark}} &  \textcolor{applegreen}{\textbf{\checkmark}}&  &
  &\textcolor{applegreen}{\textbf{\checkmark}}  & & \textcolor{applegreen}{\textbf{\checkmark}} &  \textcolor{applegreen}{\textbf{\checkmark}}&  & &  & & \textcolor{applegreen}{\textbf{\checkmark}}& && \textcolor{applegreen}{\textbf{\checkmark}} &   &\textcolor{applegreen}{\textbf{\checkmark}} \\ \hline

  \citet{shah2021nutri} &  \textcolor{applegreen}{\textbf{\checkmark}} &  &  &\textcolor{applegreen}{\textbf{\checkmark}} &  & \textcolor{applegreen}{\textbf{\checkmark}} &
  &\textcolor{applegreen}{\textbf{\checkmark}}  & & \textcolor{applegreen}{\textbf{\checkmark}} &  \textcolor{applegreen}{\textbf{\checkmark}}&  & &  & & \textcolor{applegreen}{\textbf{\checkmark}}& \textcolor{applegreen}{\textbf{\checkmark}}&&  &   &\textcolor{applegreen}{\textbf{\checkmark}} \\ \hline

    \citet{https://doi.org/10.48550/arxiv.2012.12573} &   &\textcolor{applegreen}{\textbf{\checkmark}}  &  &\textcolor{applegreen}{\textbf{\checkmark}} &  \textcolor{applegreen}{\textbf{\checkmark}}&  &
  &\textcolor{applegreen}{\textbf{\checkmark}}  & & \textcolor{applegreen}{\textbf{\checkmark}} &  \textcolor{applegreen}{\textbf{\checkmark}}&  & &  & & \textcolor{applegreen}{\textbf{\checkmark}}& &&  \textcolor{applegreen}{\textbf{\checkmark}}&   &\textcolor{applegreen}{\textbf{\checkmark}} \\ \hline

      \citet{deyoung-etal-2021-ms} &   &\textcolor{applegreen}{\textbf{\checkmark}}  &  &\textcolor{applegreen}{\textbf{\checkmark}} &  &\textcolor{applegreen}{\textbf{\checkmark}}  &
  &\textcolor{applegreen}{\textbf{\checkmark}}  & & \textcolor{applegreen}{\textbf{\checkmark}} &  \textcolor{applegreen}{\textbf{\checkmark}}&  & &  & & \textcolor{applegreen}{\textbf{\checkmark}}& &&  \textcolor{applegreen}{\textbf{\checkmark}}&   &\textcolor{applegreen}{\textbf{\checkmark}} \\ \hline

        \citet{https://doi.org/10.48550/arxiv.2110.11870} &   &\textcolor{applegreen}{\textbf{\checkmark}}  &  &\textcolor{applegreen}{\textbf{\checkmark}} &  \textcolor{applegreen}{\textbf{\checkmark}}&  &
  &\textcolor{applegreen}{\textbf{\checkmark}}  & & \textcolor{applegreen}{\textbf{\checkmark}} &  \textcolor{applegreen}{\textbf{\checkmark}}&  & &  & & \textcolor{applegreen}{\textbf{\checkmark}}& &&  \textcolor{applegreen}{\textbf{\checkmark}}&   &\textcolor{applegreen}{\textbf{\checkmark}} \\ \hline

\citet{du2020biomedical} &   &\textcolor{applegreen}{\textbf{\checkmark}}  &  &\textcolor{applegreen}{\textbf{\checkmark}} &  \textcolor{applegreen}{\textbf{\checkmark}}&  &
  \textcolor{applegreen}{\textbf{\checkmark}}&  & & \textcolor{applegreen}{\textbf{\checkmark}} &  &\textcolor{applegreen}{\textbf{\checkmark}}  & &  & & \textcolor{applegreen}{\textbf{\checkmark}}& &&  \textcolor{applegreen}{\textbf{\checkmark}}&   \textcolor{applegreen}{\textbf{\checkmark}}& \\ \hline

\citet{kedzie2018content} &   &\textcolor{applegreen}{\textbf{\checkmark}}  &  &\textcolor{applegreen}{\textbf{\checkmark}} &  \textcolor{applegreen}{\textbf{\checkmark}}&  &
  \textcolor{applegreen}{\textbf{\checkmark}}&  & & \textcolor{applegreen}{\textbf{\checkmark}} &  &\textcolor{applegreen}{\textbf{\checkmark}}  & &  & & \textcolor{applegreen}{\textbf{\checkmark}}& &\textcolor{applegreen}{\textbf{\checkmark}}&  &   &\textcolor{applegreen}{\textbf{\checkmark}} \\ \hline

\citet{Sarkar2009UsingDK} &   &\textcolor{applegreen}{\textbf{\checkmark}}  &  & \textcolor{applegreen}{\textbf{\checkmark}}&  \textcolor{applegreen}{\textbf{\checkmark}}&  &
  \textcolor{applegreen}{\textbf{\checkmark}} & &  &\textcolor{applegreen}{\textbf{\checkmark}} &  &\textcolor{applegreen}{\textbf{\checkmark}}  & & \textcolor{applegreen}{\textbf{\checkmark}} & & & &&\textcolor{applegreen}{\textbf{\checkmark}}  &    \textcolor{applegreen}{\textbf{\checkmark}}&\\ 
  \hline
  \citet{krishna-etal-2021-generating} &   &\textcolor{applegreen}{\textbf{\checkmark}}  &  &\textcolor{applegreen}{\textbf{\checkmark}} &  \textcolor{applegreen}{\textbf{\checkmark}}&  &
  &\textcolor{applegreen}{\textbf{\checkmark}}  & \textcolor{applegreen}{\textbf{\checkmark}} & &  &  & \textcolor{applegreen}{\textbf{\checkmark}}&  & & \textcolor{applegreen}{\textbf{\checkmark}}& &\textcolor{applegreen}{\textbf{\checkmark}}&  &   &\textcolor{applegreen}{\textbf{\checkmark}} \\ \hline
  
    \citet{enarvi2020generating} &   &\textcolor{applegreen}{\textbf{\checkmark}}  &  &\textcolor{applegreen}{\textbf{\checkmark}} &  \textcolor{applegreen}{\textbf{\checkmark}}&  &
  &\textcolor{applegreen}{\textbf{\checkmark}}  &  &\textcolor{applegreen}{\textbf{\checkmark}} & \textcolor{applegreen}{\textbf{\checkmark}} &  & &  & & \textcolor{applegreen}{\textbf{\checkmark}}& &\textcolor{applegreen}{\textbf{\checkmark}}&  &   &\textcolor{applegreen}{\textbf{\checkmark}} \\ \hline

      \citet{chintagunta-etal-2021-medically} &   &\textcolor{applegreen}{\textbf{\checkmark}}  &  &\textcolor{applegreen}{\textbf{\checkmark}} &  \textcolor{applegreen}{\textbf{\checkmark}}&  &
  &\textcolor{applegreen}{\textbf{\checkmark}}  &  &\textcolor{applegreen}{\textbf{\checkmark}} & \textcolor{applegreen}{\textbf{\checkmark}} &  & &  & & \textcolor{applegreen}{\textbf{\checkmark}}& &&\textcolor{applegreen}{\textbf{\checkmark}}  &   &\textcolor{applegreen}{\textbf{\checkmark}} \\ \hline

        \citet{song-etal-2020-summarizing} &   &\textcolor{applegreen}{\textbf{\checkmark}}  &  &\textcolor{applegreen}{\textbf{\checkmark}} &  \textcolor{applegreen}{\textbf{\checkmark}}&  &
  \textcolor{applegreen}{\textbf{\checkmark}}&  &  &\textcolor{applegreen}{\textbf{\checkmark}} &  &\textcolor{applegreen}{\textbf{\checkmark}}  & &  & & \textcolor{applegreen}{\textbf{\checkmark}}& &\textcolor{applegreen}{\textbf{\checkmark}}&  &   &\textcolor{applegreen}{\textbf{\checkmark}} \\ \hline

          \citet{Liu2019TopicAwarePN} &   &\textcolor{applegreen}{\textbf{\checkmark}}  &  &\textcolor{applegreen}{\textbf{\checkmark}} &  \textcolor{applegreen}{\textbf{\checkmark}}&  &
  &\textcolor{applegreen}{\textbf{\checkmark}}  &  \textcolor{applegreen}{\textbf{\checkmark}}& &  \textcolor{applegreen}{\textbf{\checkmark}}&  & &  & & \textcolor{applegreen}{\textbf{\checkmark}}& &\textcolor{applegreen}{\textbf{\checkmark}}&  &   &\textcolor{applegreen}{\textbf{\checkmark}} \\ \hline

            \citet{joshi-etal-2020-dr} &   &\textcolor{applegreen}{\textbf{\checkmark}}  &  &\textcolor{applegreen}{\textbf{\checkmark}} &  \textcolor{applegreen}{\textbf{\checkmark}}&  &
  &\textcolor{applegreen}{\textbf{\checkmark}}  &  \textcolor{applegreen}{\textbf{\checkmark}}& &  \textcolor{applegreen}{\textbf{\checkmark}}&  & &  & & \textcolor{applegreen}{\textbf{\checkmark}}& &&\textcolor{applegreen}{\textbf{\checkmark}}  &   \textcolor{applegreen}{\textbf{\checkmark}}& \\ 
  \hline

              \citet{https://doi.org/10.48550/arxiv.2109.12174} &   &\textcolor{applegreen}{\textbf{\checkmark}}  &  &\textcolor{applegreen}{\textbf{\checkmark}} &  \textcolor{applegreen}{\textbf{\checkmark}}&  &
  &\textcolor{applegreen}{\textbf{\checkmark}}  &  &\textcolor{applegreen}{\textbf{\checkmark}} &  &  &\textcolor{applegreen}{\textbf{\checkmark}} &  & & \textcolor{applegreen}{\textbf{\checkmark}}& &&\textcolor{applegreen}{\textbf{\checkmark}}  &   &\textcolor{applegreen}{\textbf{\checkmark}} \\ 
  \hline

  \citet{zhang2018learning} & \textcolor{applegreen}{\textbf{\checkmark}}  &  &  &\textcolor{applegreen}{\textbf{\checkmark}} &  \textcolor{applegreen}{\textbf{\checkmark}}&  &
  &\textcolor{applegreen}{\textbf{\checkmark}}  &  &\textcolor{applegreen}{\textbf{\checkmark}} &  \textcolor{applegreen}{\textbf{\checkmark}}&  & &  & & \textcolor{applegreen}{\textbf{\checkmark}}& &\textcolor{applegreen}{\textbf{\checkmark}}&  &   &\textcolor{applegreen}{\textbf{\checkmark}} \\ 
  \hline

\citet{aramaki-etal-2009-text2table} &   &\textcolor{applegreen}{\textbf{\checkmark}}  &  & \textcolor{applegreen}{\textbf{\checkmark}}&  \textcolor{applegreen}{\textbf{\checkmark}}&  &
  \textcolor{applegreen}{\textbf{\checkmark}} & &  &\textcolor{applegreen}{\textbf{\checkmark}} &  &\textcolor{applegreen}{\textbf{\checkmark}}  & &  &\textcolor{applegreen}{\textbf{\checkmark}} & & &&\textcolor{applegreen}{\textbf{\checkmark}}  &    &\textcolor{applegreen}{\textbf{\checkmark}}\\ 
  \hline

    \citet{macavaney2019ontology} & \textcolor{applegreen}{\textbf{\checkmark}}  &  &  &\textcolor{applegreen}{\textbf{\checkmark}} &  \textcolor{applegreen}{\textbf{\checkmark}}&  &
  &\textcolor{applegreen}{\textbf{\checkmark}}  &  &\textcolor{applegreen}{\textbf{\checkmark}} &  \textcolor{applegreen}{\textbf{\checkmark}}&  & &  & & \textcolor{applegreen}{\textbf{\checkmark}}& &&\textcolor{applegreen}{\textbf{\checkmark}}  &   \textcolor{applegreen}{\textbf{\checkmark}}& \\ 
  \hline

    \citet{zhang2019optimizing} & \textcolor{applegreen}{\textbf{\checkmark}}  &  &  &\textcolor{applegreen}{\textbf{\checkmark}} &  \textcolor{applegreen}{\textbf{\checkmark}}&  &
  &\textcolor{applegreen}{\textbf{\checkmark}}  &  &\textcolor{applegreen}{\textbf{\checkmark}} &  \textcolor{applegreen}{\textbf{\checkmark}}&  & &  & & \textcolor{applegreen}{\textbf{\checkmark}}& \textcolor{applegreen}{\textbf{\checkmark}}&&  &   &\textcolor{applegreen}{\textbf{\checkmark}} \\ 
  \hline
   \citet{pre} &   &\textcolor{applegreen}{\textbf{\checkmark}}  &  & \textcolor{applegreen}{\textbf{\checkmark}}&  \textcolor{applegreen}{\textbf{\checkmark}}&  &
  \textcolor{applegreen}{\textbf{\checkmark}} & &  &\textcolor{applegreen}{\textbf{\checkmark}} &  &\textcolor{applegreen}{\textbf{\checkmark}}  & & \textcolor{applegreen}{\textbf{\checkmark}} & & & &&\textcolor{applegreen}{\textbf{\checkmark}}  &    &\textcolor{applegreen}{\textbf{\checkmark}}\\ 
  \hline
        
    \citet{hu2021word} & \textcolor{applegreen}{\textbf{\checkmark}}  &  &  &\textcolor{applegreen}{\textbf{\checkmark}} &  \textcolor{applegreen}{\textbf{\checkmark}}&  &
  &\textcolor{applegreen}{\textbf{\checkmark}}  &  &\textcolor{applegreen}{\textbf{\checkmark}} &  \textcolor{applegreen}{\textbf{\checkmark}}&  & &  & & \textcolor{applegreen}{\textbf{\checkmark}}& \textcolor{applegreen}{\textbf{\checkmark}}&&  &   \textcolor{applegreen}{\textbf{\checkmark}} &\\ 
  \hline
  
      \citet{kondadadi2021optum} & \textcolor{applegreen}{\textbf{\checkmark}}  &  &  &\textcolor{applegreen}{\textbf{\checkmark}} &  \textcolor{applegreen}{\textbf{\checkmark}}&  &
  &\textcolor{applegreen}{\textbf{\checkmark}}  &  &\textcolor{applegreen}{\textbf{\checkmark}} &  \textcolor{applegreen}{\textbf{\checkmark}}&  & &  & & \textcolor{applegreen}{\textbf{\checkmark}}& &&\textcolor{applegreen}{\textbf{\checkmark}}  &    &\textcolor{applegreen}{\textbf{\checkmark}}\\ 
  \hline
  
        \citet{mrini2021gradually} &   &\textcolor{applegreen}{\textbf{\checkmark}}  &  \textcolor{applegreen}{\textbf{\checkmark}}& &  \textcolor{applegreen}{\textbf{\checkmark}}&  &
  &\textcolor{applegreen}{\textbf{\checkmark}}  &  &\textcolor{applegreen}{\textbf{\checkmark}} &  \textcolor{applegreen}{\textbf{\checkmark}}&  & &  & & \textcolor{applegreen}{\textbf{\checkmark}}& &&\textcolor{applegreen}{\textbf{\checkmark}}  &    &\textcolor{applegreen}{\textbf{\checkmark}}\\ 
  \hline
          \citet{Abacha2019OnTS} &   &\textcolor{applegreen}{\textbf{\checkmark}}  &  \textcolor{applegreen}{\textbf{\checkmark}}& &  \textcolor{applegreen}{\textbf{\checkmark}}&  &
  &\textcolor{applegreen}{\textbf{\checkmark}}  &  &\textcolor{applegreen}{\textbf{\checkmark}} &  \textcolor{applegreen}{\textbf{\checkmark}}&  & &  & & \textcolor{applegreen}{\textbf{\checkmark}}& &&\textcolor{applegreen}{\textbf{\checkmark}}  &    &\textcolor{applegreen}{\textbf{\checkmark}}\\ 
  \hline
  
            \citet{yadav2022question} &   &\textcolor{applegreen}{\textbf{\checkmark}}  &  \textcolor{applegreen}{\textbf{\checkmark}}& &  \textcolor{applegreen}{\textbf{\checkmark}}&  &
  &\textcolor{applegreen}{\textbf{\checkmark}}  &  &\textcolor{applegreen}{\textbf{\checkmark}} &  \textcolor{applegreen}{\textbf{\checkmark}}&  & &  & & \textcolor{applegreen}{\textbf{\checkmark}}& &&\textcolor{applegreen}{\textbf{\checkmark}}  &    \textcolor{applegreen}{\textbf{\checkmark}}&\\ 
  \hline

 \end{tabular}

}
\end{table}

\textbf{Learning process: } Most of the existing techniques to solve the task of MDS can fall under these three categories: (1) Rule-based learning process \cite{Sarkar2009UsingDK,pre}: This involves the use of sentence ranking, sentence extraction, and clustering techniques. Most of these works are pre-deep learning era (before 2010), (2) Machine learning based \cite{aramaki-etal-2009-text2table, sarkar2011using}: This involves the use of classical machine learning techniques such as supervised clustering techniques, and (3) Deep learning \cite{https://doi.org/10.48550/arxiv.2006.01997, cai2022covidsum,https://doi.org/10.48550/arxiv.2007.03405,Abacha2019OnTS, CHQ, Mrini2021JointSO, zeng-etal-2020-meddialog, joshi-etal-2020-dr, delbrouck-etal-2021-qiai}: With the advent of neural networks and transformers, most of the current MDS approaches leverage these techniques. Table \ref{tab:DL} lists the works using different types of deep learning techniques.
\par
\textbf{Notion of importance: } Unlike other domains, medical domain is a very high stake area that requires certain summarization models to give more focus or weightage to certain aspects of the summary. Thus, the most significant distinction between the existing work would be the notion of importance used to generate the final summary. A diverse set of objectives ranging from consistency \cite{zhang2019optimizing}, copy mechanism \cite{zhang2018learning}, and more weightage to extraction \cite{https://doi.org/10.48550/arxiv.2101.04840} have been explored in an attempt to solve the MDS task in an efficient manner.
\par
\textbf{Use of Knowledge Base: } External Knowledge bases (KBs) can improve natural language processing (NLP) in a number of ways \cite{zouhar2022artefact} especially in the medical domain as medical information uses very specific terminologies and vocabulary than general information. Based on this, we can classify current MDS works on the basis of whether they use a KB \cite{macavaney2019ontology,sotudeh2020attend, hu2021word,cai2022covidsum} or not \cite{https://doi.org/10.48550/arxiv.2104.13498, https://doi.org/10.48550/arxiv.2105.00816, delbrouck-etal-2021-qiai, zhang2019optimizing}. If they use KB, what type of KB they are using is crucial as there can be multiple ways in which a KB can be used to improve performance such as using a medical database directly as an external knowledge source \cite{macavaney2019ontology,sotudeh2020attend} or creating a knowledge graph from a dataset to create a local knowledge base \cite{hu2021word,cai2022covidsum}. A detailed list of works that use knowledge bases networks can be found
at Table \ref{tab:KB}. 
\par

\begin{table}[tp]  \centering
\scriptsize
\caption{\textbf{Comprehensive list of works that use specific Knowledge bases to generate output summary.}}\label{tab:KB}
\renewcommand{\arraystretch}{2}
\begin{tabular}{|p{0.2\textwidth}|p{0.6\textwidth}|}
\hline
\textbf{Knowledge Bases (KB)} & \textbf{Works using this KB}\\
\hline
Pre-trained medical Transformer as base model & \citet{https://doi.org/10.48550/arxiv.2008.11293}, \citet{du2020biomedical}, \citet{yadav2022question}
\\\hline

Medical Database such as QuickUMLS \cite{Soldaini2016QuickUMLSAF}  & \citet{zhang2018learning}, \citet{macavaney2019ontology}\\\hline

Knowledge Graphs & \citet{hu2021word}, \citet{cai2022covidsum} \\\hline

\end{tabular}
\end{table}

\section{Evaluation Techniques} \label{sec:eval}
There have been a lot of diverse medical input documents and numerous attempts in solving the medical task summarization task. However, the evaluation of these summaries is still an open problem. Most of the existing works use standard n-gram matching metrics, (Recall-Oriented Understudy for Gisting Evaluation) ROUGE \cite{lin-2004-rouge}, Metric for Evaluation of Translation with Explicit ORdering (METEOR) \cite{banerjee-lavie-2005-meteor}, and Bilingual Evaluation Understudy (BLEU) \cite{papineni2002bleu} as evaluation measures to evaluate the quality of generated summaries. A few works also uses semantic evaluation metrics such as Sentence Mover Similarity (SMS) \cite{clark-etal-2019-sentence} and BERTScore \cite{zhang2019bertscore}. The most popular way of measuring the quality of text summarization in the field is the Recall-Oriented Understudy for Gisting Evaluation (ROUGE), which is largely reliant on the similarity between the generated summary and the reference summary in terms of n-gram overlap. However, a major downside to ROUGE is its need for a reference summary of good quality. This was highlighted in many studies \cite{https://doi.org/10.48550/arxiv.2007.12626,10.1007/978-3-030-99736-6_21}, that pointed out that some reference summaries are not of consistent and satisfactory quality.\par
However, the best way to evaluate the quality of a generated summary is to perform extensive human evaluations. The standard predefined metrics \cite{10.1016/j.jbi.2022.103999, https://doi.org/10.48550/arxiv.2008.11293, shah-etal-2021-nutri} to measure in the human evaluation process are (1) \textit{Informativeness or Relevance: }It measures how many important and relevant facts and information summary are able to retain, (2) \textit{Coherence: }measures that summary sentences or paragraphs have a smooth logical transition, (3) \textit{Redundancy: } It measures that summary should not contain repeated information or facts, (4) \textit{Fluency: } It measures the grammatical correctness of summary, (5) \textit{Consistency or Factuality: }It measures the factual correctness of summary with respect to the source article, and (6) \textit{Contradiction: } It measures whether there is some information present in the summaries that contradict some other information or at disagreement with another piece of information. Moramarco et al. \cite{https://doi.org/10.48550/arxiv.2104.04412} conduct a human evaluation of the quality of Clinical SOAP notes generated by state-of-the-art (SOTA) text summarization models (Both extractive and abstractive models.) by asking evaluators to count medical facts, and computing precision, recall, F-score, and accuracy from those raw counts. One fascinating observation from this study was that both abstractive and extractive models don’t hallucinate any medical facts as abstractive models also tend to copy many phrases from the source text. However, they observed that the only hallucination that the abstractive models do in those samples is a numerical one showing the limitations of current SOTA models to comprehend numerical relations. This limited numerical literacy of SOTA language models is also highlighted in many previous studies \cite{thawani-etal-2021-representing}. However, human evaluations are very expensive and time-consuming, thus not scalable for large models and datasets. \par
\begin{figure}[t]
\includegraphics[width=.9\textwidth]{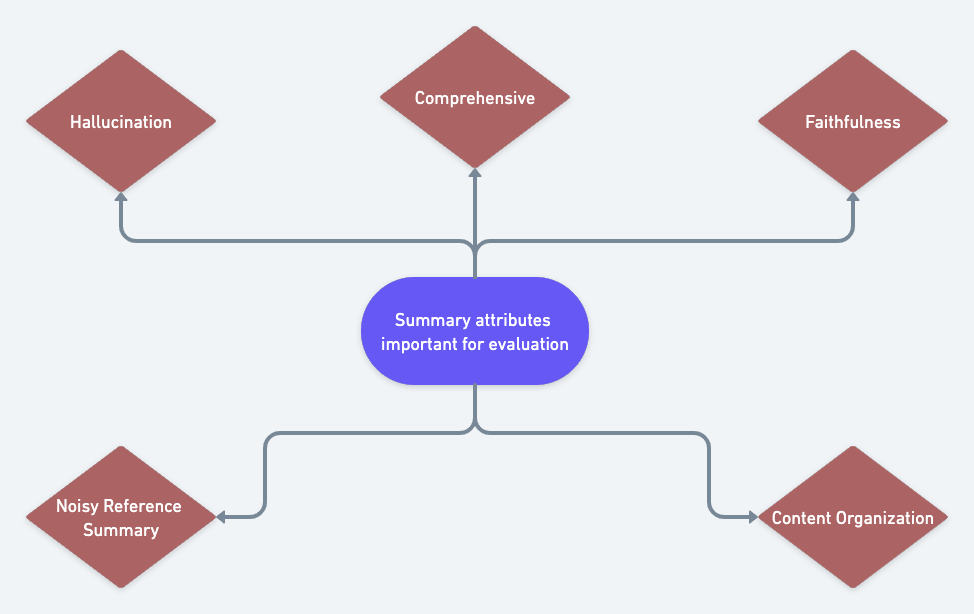}\centering
\caption{Visual representation of specific attributes to be considered while doing evaluation of medical summaries} \label{evalpic}
\end{figure}

There are certain attributes that are specific to medical documents which above mentioned metrics fail to measure (Figure \ref{evalpic}). Following are the unique traits that one must consider while dealing with medical documents: (1) Comprehensive summaries: Adams et al. \cite{https://doi.org/10.48550/arxiv.2104.13498} study that medical summaries are densely packed with medical terminologies and entities. They found out that 20.9\% of
the words in a summary is medical entities as per UMLS \cite{UMLS} dataset as compared to 14.1\% in the source document. They also showed that summaries contain 26 unique medical entities on average whereas  source document contains 265 unique medical entities. This is a compression ratio of only 10 as compared to the compression ratio of 45 for all tokens. Goel et al. \cite{https://doi.org/10.48550/arxiv.2105.00816} highlight that performance degrades when the number of entities increases in the CNN/DailyMail dataset. This necessitates the need for domain-specific fact-based evaluation approaches which can encode a deeper knowledge of clinical concepts and their complex semantic and temporal relations to assess the quality of generated medical entities. (2) Hallucination: Hallucination is one of the leading limitations of existing summarization systems. In the case of clinical and medical settings, hallucinations in summaries can lead to serious treatment misjudgment and errors. Wallace et al. \cite{https://doi.org/10.48550/arxiv.2008.11293} also highlighted that metrics like ROUGE fail to capture hallucinations in summaries. Some studies \cite{https://doi.org/10.48550/arxiv.2103.11921,https://doi.org/10.48550/arxiv.2109.12174} also showed this issue of hallucinated content and medically incorrect entities and outputs in generated summaries through human evaluation. (3) Faithfulness: Shing et al. \cite{https://doi.org/10.48550/arxiv.2101.04840} define faithful summary as a summary that is faithful to the source (doesn't contain any information from outside of the source and whatever present should be factually correct) and relevant as measured by the reference summary (contains all the key and salient information). (4) Content Organization: Content organization refers to the structure of a document and the best metric to measure this is Coherence \cite{https://doi.org/10.48550/arxiv.2007.12626}. Barzilay et al. \cite{barzilay-lapata-2005-modeling} define coherence as \textit{information about the same entities are considered to be more coherent than information with the abrupt transition of topics or entities}. However, Adams et al. \cite{https://doi.org/10.48550/arxiv.2104.13498} showed that CLINSUM (dataset of health record summarization) has a swift and abrupt topic and entity transitions with only 34\% of the entities repeating. They also showed that ROUGE is insufficient to capture the coherence in the CLINSUM dataset using a pairwise ranking approach \cite{barzilay-lapata-2005-modeling}. This problem warrants the need of developing domain-specific models of coherence, which can handle these abrupt topic shifts, and able to capture these relations between different medical entities present in the summaries. Some studies \cite{https://doi.org/10.48550/arxiv.2007.12626, https://doi.org/10.48550/arxiv.2007.12626, bhandari-etal-2020-evaluating} also showed the importance and need of domain-oriented evaluation metrics. (5) Noisy Reference Summaries: Dependency on a good quality reference summary is one of the major limitations of metrics like ROUGE or BLEU \cite{kryscinski-etal-2019-neural,https://doi.org/10.48550/arxiv.2007.12626} as reference summary itself can be of below-par quality. Kripalani et al. \cite{BP} discovers that discharge summaries often miss important and salient information such as diagnostic results (missing from 33\%-
63\% time), treatment course of a patient(7\%-22\%), discharge medications (2\%-40\%), test
results (65\%), counseling (90\%-92\%), and follow-up and future steps (2\%-43\%). This shows that complete dependency on reference summary for evaluation is not ideal in clinical settings as reference summary in itself lacks critical information.\par

From the above discussion, it is evident that we need new evaluation metrics that are capable of measuring all these aspects of a summary of a medical document. Recently various attempts have been made by the community to propose such efficient and domain-specific metrics. Some of the contributions are as follow: (1) KG metrics \cite{shah-etal-2021-nutri}: Shah et al. \cite{shah-etal-2021-nutri} propose two metrics KG(G) and KG(I) to capture relevance and faithfulness of generated summary respectively. Both these metrics are based on entity and relation matching. 
KG(G) captures relevance by calculating the number of overlapping entity-relation-entity pairs in the generated summary and reference summary. Similarly, KG(I) calculates the number of overlapping entity-relation-entity pairs in the generated summary and input source document to measure the faithfulness of the summary with respect to the input document. To calculate the entity-relation-entity pairs for generated summaries, gold summaries, and input documents, they run an entity tagging model and relation classification model on all these three documents. Then they match the gold summaries
$(e^G_i, e^G_j, r^G)$ pairs and input document $(e^I_i, e^I_j, r^I)$ pairs using cosine similarity with the  generated output summary $(e^o_i, e^o_j, r^o)$ pairs. They consider all those pairs to be a match that has a cosine score of more than a fixed threshold of 0.7. (2) Aggregation Cognisance (Ag) metric: Shah et al. \cite{shah-etal-2021-nutri} proposes a metric Aggregation Cognisance (Ag) to capture the capability of the model to generate output summary that is aware of the right aggregation or entailment (contradiction or agreement) from the input source document by using a classifier to measure and compare the entailment in the generated output summary and the input source document. (3) Diversity metric \cite{https://doi.org/10.48550/arxiv.1904.02281, shah-etal-2021-nutri}: Rao et al. \cite{https://doi.org/10.48550/arxiv.1904.02281}  proposed a metric to measure diversity in model's outputs by computing the ratio of unique trigrams present in a generated summary. (4) Readability evaluation metrics \cite{https://doi.org/10.48550/arxiv.2012.12573}: Guo et al. uses Flesch-Kincaid grade level \cite{kincaid1975derivation}, Gunning fog index \cite{flory1992measuring},
and Coleman-Liau index \cite{coleman1975computer} to compute the ease of readability and fluency of generated summaries. (5) Fact-based Evaluation: To measure the accuracy of medical facts generated by the model in output summary, Enarvi et al. \cite{enarvi2020generating} utilizes a machine-learning clinical fact
extractor module that is capable of extracting medical facts such as treatment or diagnosis and fine-grained attributes such as body part or medications. They use this model to extract facts from the generated summary and reference summary and calculate the F1 score for both extracted facts. Similar type of factual correctness metrics has been used in several other works \cite{hu2021word,zhang2019optimizing,chintagunta2021medically}. Enarvi et al. \cite{enarvi2020generating} used Negex metric \cite{harkema2009context} to capture the capability of model to evaluate the negative status of medical concepts by computing the negations in generated summary and calculating whether the negations were accurate for the medical facts present in the generated summary. (6) Faithfulness and Hallucination Metrics: Shing et al. \cite{https://doi.org/10.48550/arxiv.2104.13498} proposes two evaluation metrics, Faithfulness-adjusted Precision and Incorrect Hallucination Rate to measure faithfulness and hallucination in generated summaries. Faithfulness-adjusted Precision is defined as:
\begin{equation}
    Faithfulness adjusted Precision =  \frac{(O\cap\ G \cap S)}{O}
\end{equation}
where $O$ refers to generated output summary, $G$ refers to gold reference summary, and $S$ refers to input source document. In a similar fashion, Incorrect Hallucination Rate is calculated as:
\begin{equation}
    Incorrect Hallucination Rate = \frac{(O-(O \cap G)-(O \cap S))}{O}
\end{equation}\par
Overall, the discussed metrics have still their limitations; however, there is a great scope for future improvement in the area of evaluation techniques for medical summaries. We provided a comprehensive study of these evaluation metrics with their pros and cons in Table \ref{tab:evaluation}. 

\vspace{-1em}
\section{Ehtical Consideration}\label{EC}
Deep learning is a powerful tool that can be used for good or bad. On the one hand, deep learning can be used to create amazing new technology that can be used to improve our lives. On the other hand, deep learning can also be used for malicious purposes, such as creating deepfakes or propaganda bots. As with any technology, there are ethical considerations to be taken into account when using deep learning. Some of the ethical considerations of deep learning include data privacy, data bias, and the impact of artificial intelligence on society. The ethical implications of clinical deep learning are still being explored.
However, there are some potential concerns that have been raised.\par
\begin{table}[H]  \centering 
\scriptsize
\caption{\textbf{Comparative study of evaluation techniques for  Medical Document Summarization.}}\label{tab:evaluation}
\renewcommand{\arraystretch}{2}
\begin{tabular}{|p{0.3\textwidth}| p{0.6\textwidth}|}
\hline
\textbf{Metric name \& corresponding paper} & \textbf{Pros \& Cons} \\
\hline

\multirow{5}{0.2\textwidth}{KG metrics. \citet{shah-etal-2021-nutri}} & \textbf{Advantages} \\
& - Designed to capture relevance and faithfulness of generated summary with respect to reference summary and the input source document. \\
& \textbf{Disadvantages} \\
& - dependent on a good quality reference summary and Performance may vary with the use of different entity tagging and relation classification models \\
\hline

\multirow{5}{0.2\textwidth}{Aggregation Cognisance (Ag) metric. Shah et al. \cite{shah-etal-2021-nutri}} & \textbf{Advantages} \\
& - Designed to capture the capability of a model to generate an output summary that is aware of the right
entailment (contradiction or agreement) with respect to input source. \\
& \textbf{Disadvantages} \\
& - Performance may vary with the use of different entailment classifiers. \\

\hline

\multirow{5}{0.2\textwidth}{Diversity metric. \citet{https://doi.org/10.48550/arxiv.1904.02281, shah-etal-2021-nutri}} & \textbf{Advantages} \\
& - Aims to Capture the diversity in generated output summary \\
& \textbf{Disadvantages} \\
& - The technique is very naive as it considers only n-gram overlaps.\\

\hline
\multirow{5}{0.2\textwidth}{Readability evaluation metrics. \citet{https://doi.org/10.48550/arxiv.2012.12573}} & \textbf{Advantages} \\
& - Designed to measure the readability and fluency of generated output. \\
& \textbf{Disadvantages} \\
& - The technique may not be robust as it uses rule-based metrics. \\

\hline
\multirow{5}{0.2\textwidth}{Fact-based Evaluation. Enarvi et al. \cite{enarvi2020generating}} & \textbf{Advantages} \\
& - Aims to measure the factual correctness of medical facts present in generated summary. \\
& \textbf{Disadvantages} \\
& - Different works use different fact extractor leading to a lot of inconsistencies among results. \\

\hline
\multirow{5}{0.2\textwidth}{Faithfulness and Hallucination Metrics. Shing et al. \cite{https://doi.org/10.48550/arxiv.2104.13498}} & \textbf{Advantages} \\
& - Aims to measure faithfulness of summaries and detection of hallucination in summaries. \\
& \textbf{Disadvantages} \\
& - The technique is very naive as it considers only syntactic overlap. \\

\hline
\end{tabular}
\end{table}
\textbf{Bias and Fairness: }Machine learning is a data-driven field that makes the models and algorithms susceptible to the social bias present in the datasets. This will make the models to make decisions that are skewed against certain communities of the society \cite{garrido2021survey}. Chen et al. \cite{chen2019can} conducted two studies to understand the bias in clinical deep learning models, one for intensive care unit (ICU) mortality prediction and second for 30-day psychiatric readmission prediction. They reported that these models underperform on both tasks with respect to 
women, ethnic and racial minorities. Zhang et al. \cite{https://doi.org/10.48550/arxiv.2003.11515} also studied the societal bias in language models and found that they too are susceptible to biases by recommending hospitals and clinics to white people and prison to black people which is very worrisome sign and raises a question mark on real world application of such systems. This warrants the need of debiased and fair machine learning techniques \cite{https://doi.org/10.48550/arxiv.2111.08711, https://doi.org/10.48550/arxiv.2203.09860, DB}. Similar types of studies about bais and debiasing techniques must be applied to medical summarization systems too before deploying them in real-world settings. 

\par
\textbf{Data Privacy: } The digitisation of medical records is underway for the convenience of patients, doctors, and medical staff. The digitization of medical documents is a key part of medical care and is also of great importance in improving medical care. The digitization of medical documents has many benefits. It helps to save time and paper and to improve the quality of medical care. It also helps to reduce medical errors. The digitization of medical documents is also of great importance in terms of both medical research and clinical process. The digitization of information has led to a number of privacy concerns. The main concern is that digitization makes it easier for organizations to collect, store and use personal data which makes the patient's data susceptible to various security attacks 
\cite{priya2017survey,kaye2012tension}. This requires the need for data anonymization which is the process of de-identifying data so that it can no longer be traced back to an individual; this can be done by removing or encrypting personal identifiers such as names, addresses, and social security numbers \cite{olatunji2022review}. Data anonymization is a critical process that must be done before creating or releasing any dataset for the MDS task.

\section{Discussion} \label{Dis}
Medical Document summarization has been researched quite frequently in recent years with multiple diverse models. We discuss how medical documents can be categorized into fine-grained categories with each type presenting its unique challenges to deal with. It can be seen that some documents (Research articles, Radiology reports, and medical dialogue) have been explored more than other medical documents (Electronic health records and consumer health questions). The reason for this can be attributed to the difficulty in obtaining the datasets for Electronic health records and Consumer Health Questions sub-tasks. We also categorize the techniques to solve Medical Document summarization to understand the current trends.
We also categorized current works on the basis of input, output, and method used. It is evident that most of the current work revolves around single documents as input, abstractive summaries as outputs, and deep learning or transformer as their base model. Some of the works also use external Knowledge bases in the form of medical databases or knowledge graphs to further improve the performance. There are also few that focus on specific medical domains such as COVID articles or Radiology reports. Recently, the community is also moving towards a hybrid of extractive and abstractive summarization approaches (Extract-then-Abstract) to improve the faithfulness of generated summaries.
After conducting a comprehensive study of evaluation metrics, we can make the following observations: (1) Standard evaluation metrics are insufficient to capture the special aspects of medical summaries. (2) There are a lot of inconsistencies in the human evaluation process as different researchers measure different aspects of the summary; thus we compile and report all these aspects that need to be considered while doing the human evaluation. (3) We also study the existing medical domain-specific metrics that the community should use alongside standard evaluation metrics. In addition to all these, we also discuss the need of ensuring that deep learning is used responsibly and with consideration for the potential impacts on individuals and society. Broadly, there are two ethical concerns that one must consider while deploying their model in the real world; Fairness: one must ensure that their algorithm is used fairly and free from bias, and data privacy: one must ensure that the data collected is kept secure and that unauthorized access is not granted.

\section{Future Work} \label{sec:future}
The MDS task is relatively new, and the work done so far has only scratched the surface of what
this field has to offer. In this section, we discuss the future scope of the MDS task, including some
possible improvements in existing works, as well as some possible new directions.
\subsection{Scope of improvement}
\textbf{Better evaluation metrics: } Most of the existing works still use standard evaluation techniques like ROUGE scores and BLEU scores to measure the quality of the generated medical summary. However, these metrics suffer from a lot of limitations. Many works try to compensate for the limitations of these metrics by conducting a human evaluation procedure. But these are very expensive and time-consuming processes making them unscalable for larger datasets. Many attempts have been made to formulate evaluation metrics that are capable of measuring medical summaries specific aspects such as readability, faithfulness, fact-based evaluation, etc. However, all these metrics suffer from their own limitations making a room for the scope of improvement which are discussed as follows: (1) All of these metrics work at a syntactic level which makes them unsuitable to capture semantic overlaps. (2) Most of these metrics are over-dependent on reference summary. (3) There are no metrics that are designed to work in multimodal settings. (4) There is also a need for metrics that are capable of measuring the correctness of the relationship between different generated medical concepts by designing medical domain-aware metrics. We need to overcome all these limitations in order to improve overall medical summarization systems.
\par
\textbf{Multimodality: }Generating multi-modal summaries in the medical domain is an unsolved problem. With medical history summarization systems, generating relevant text summaries accompanied by relevant sonograms, cardiograms, or other relevant visual/aural content is an intuitive next step. There have been several unsupervised efforts to generate multi-modal summaries \cite{jangra2020text, 10.1145/3397271.3401232, 10.1145/3404835.3462877}; however, the community still lacks relevant datasets and models to obtain supervised multi-modal summarization systems for the medical domain.
\par
\textbf{Faithful summaries:} One open challenge in the current summarization system is the  unfaithfulness of generated summaries with respect to the input document \cite{https://doi.org/10.48550/arxiv.2210.01877, https://doi.org/10.48550/arxiv.2005.00661}. Solving this problem of unfaithful and inconsistent summaries becomes of utmost priority when working in the medical domain as a minor error can lead to a wrong judgment or decision. Due to this reason, the community either depends on extractive summarization techniques or uses the extract-then-abstract \cite{https://doi.org/10.48550/arxiv.2104.13498} methodology to create an evidence fallback system to check the validity of generated summary. Hence the community still lacks an abstractive summarization model that is capable of generating faithful and factually consistent summaries.
\subsection{New directions}
\textbf{Focus on Numerical Literacy: } Moramarco et al. \cite{https://doi.org/10.48550/arxiv.2104.04412} showed that abstractive models tend to numerical errors while generating summaries showing the limitations of current SOTA models to comprehend numerical values and relations. Thawani et al. \cite{thawani-etal-2021-numeracy, thawani-etal-2021-representing} also discussed how numbers are often neglected in natural language processing and thus many SOTA language models have limited  numerical
literacy. Numerical values are one of the salient features of medical documents such as patient's body vital information or lab results \cite{num}. Thus developing summarization systems that are capable of numerical reasoning is of primary importance.

\par
\textbf{GPT-3 based data annotation:} Obtaining datasets of medical documents is one of the biggest challenge faced by the community restricting the researchers to only a few datasets available online. The two major reasons for this are the privacy of patient information and if the documents are anonymized, expensive manual annotation of those documents as this can be done only by medical professionals. Chintagunta et al. \cite{chintagunta-etal-2021-medically} showed how GPT-3 \cite{https://doi.org/10.48550/arxiv.2005.14165} can be used as a tool to generate a dataset for medical dialogue summarization task. They reported that they can obtain the same performance as that of a human-labeled dataset with a 30x smaller amount of human-labeled. GPT-3 is a few-shot learner language model; thus showing only a few human-labeled examples, it can be used to annotate or label the rest of the data. Similar strategy can be used to develop datasets for related MDS subtasks.
\par
\textbf{Explainable summarization: }With the introduction of explainable artificial intelligence (AI) \cite{arrieta2020explainable}, it is required
to provide explanations/interpretations behind any decision taken
by a machine learning algorithm especially in the medical domain which is a high stake domain. Medical professionals need an interpretable mechanism to understand why the model generated that particular information. However, most of the existing interpretable work revolves around classification tasks \cite{https://doi.org/10.48550/arxiv.1602.04938, https://doi.org/10.48550/arxiv.1703.01365} which can not be used to interpret natural language generation models. Xu et al. \cite{https://doi.org/10.48550/arxiv.2106.01518} propose an ablation and attribution strategy to interpret a summarizer's decision at every time step. however, there is no existing research on explainable summarization in the medical domain showing a research gap between both these areas which needs to be filled.
\par
\textbf{Query-based summarization: }A lot of work has been done in query-based text summarization \cite{QMS, QMS2} which aims at extracting information that tries to answer a query about input source text. This type of use case can be very relevant to medical domain as medical professionals can leverage this feature of query based summarization by generating only relevant summary related to medical professional or patient query from 
plethora of documents. We believe that query-based summarization setup can imporve the overall adaption of MDS tasks in real-world settings.
\par
\textbf{Adversarial study of medical summarization systems: } Recent works have showed that adversarial examples can be generated by applying small
perturbations or noise to the inputs which will make deep learning models to perfrom abnormally \cite{huang2011adversarial}. The robustness of medical summarization systems for medical documents such as electronic health records is especially critical because of the high public interests of such a high stake domain that make it more susceptible to such adversarial attacks. Wang et al. \cite{wang-etal-2020-utilizing} showed the vulnerability of the existing summarization systems by constructing adversarial samples for EHR data. They also proposed a novel defense strategy to detect adversarial examples in datasets. In order to deploy such medical summarization systems, this type of testing needs to be done to test the robustness of models.

\par
\textbf{Fair and ethical summarization systems: } Despite so much research being done around Fair and Ethical NLP \cite{fort2016yes}, there is still no work discussing the fairness of medical summarization systems. Fair and ethical summarization systems are those systems that are developed with the goal of producing summary outputs that accurately and ethically reflect the original source material. These systems must be designed to ensure that all potentially sensitive or proprietary information is kept secure, and that summary outputs are not biased toward any particular point of view. 

\par
\section{Conclusion} \label{sec:conc}
The internet has drastically changed the way medical documents are created and accessed. In the past, they were often handwritten which made them hard to share and find. Now, they are typically created electronically which makes them much easier to both access and share. The internet has also allowed medical professionals to easily share documents with one another which has ultimately improved patient care and medical research. This paper provides a survey to introduce users and researchers to the techniques and current trends in the Medical Summarization task. We cover the formal definition of the Medical Summarization task, a detailed analysis of different medical tasks based on the type of medical documents and specific datasets and challenges associated with them, a detailed categorization of existing works based on input, output, and technique, and an in-depth look at the evaluation metrics utilized to measure the quality of the summaries. To finish, we suggest some potential future directions for further research. We are confident that this survey will encourage more work in medical document summarization.

\bibliographystyle{ACM-Reference-Format}
\bibliography{main}

\end{document}